# Synthetic Data Blueprint (SDB): A modular framework for the statistical, structural, and graph-based evaluation of synthetic tabular data

**Vasileios C. Pezoulas[1], Nikolaos S. Tachos[1], Eleni Georga[1], Kostas Marias[1], Manolis Tsiknakis[1], and Dimitrios I. Fotiadis[1]**

[1]SYNTHAINA AI, Greece

Corresponding author: Prof. Dimitrios I. Fotiadis, email: dimitris.fotiadis30@gmail.com.

**Abstract**

In the rapidly evolving era of Artificial Intelligence (AI), synthetic data are widely used to accelerate innovation while preserving privacy and enabling broader data accessibility. However, the evaluation of synthetic data remains fragmented across heterogeneous metrics, ad-hoc scripts, and incomplete reporting practices. To address this gap, we introduce Synthetic Data Blueprint (SDB), a modular Pythonic based library to quantitatively and visually assess the fidelity of synthetic tabular data. SDB supports: (i) automated feature-type detection, (ii) distributional and dependency-level fidelity metrics, (iii) graph- and embedding-based structure preservation scores, and (iv) a rich suite of data visualization schemas. To demonstrate the breadth, robustness, and domain-agnostic applicability of the SDB, we evaluated the framework across three real-world use cases that differ substantially in scale, feature composition, statistical complexity, and downstream analytical requirements. These include: (i) healthcare diagnostics, (ii) socioeconomic and financial modelling, and (iii) cybersecurity and network traffic analysis. These use cases reveal how SDB can address diverse data fidelity assessment challenges, varying from mixed-type clinical variables to high-cardinality categorical attributes and high-dimensional telemetry signals, while at the same time offering a consistent, transparent, and reproducible benchmarking across heterogeneous domains.

**Keywords:** Synthetic data, data fidelity assessment, graph-based evaluation, statistical similarity metrics, embedding-based metrics, structural data analysis, trustworthy AI.

## 1. Introduction

The widespread adoption of data-driven technologies across healthcare, finance, education, and other critical sectors has intensified the demand for accessible, high-quality data [1-3]. However, real-world data are often sensitive, proprietary, or restricted due to ethical considerations and regulatory frameworks, such as, the General Data Protection Regulation (GDPR) [4], the Health Insurance Portability and Accountability Act (HIPAA) [5], and emerging AI governance policies [6]. These constraints significantly limit data sharing and obscure the development, validation, and deployment of AI-based systems. As a result, synthetic data generation has emerged as a promising solution to enable data access without exposing real data. Synthetic data can support AI model training, experimentation, and benchmarking without exposing sensitive information by creating artificial records that preserve the statistical and structural properties of the real data [7].

Contemporary generative models, including Bayesian networks [8], Generative Adversarial Networks (GANs) [9], Variational Autoencoders (VAEs) [10], transformer-based architectures [11], and diffusion models [12], have accelerated the interest in synthetic data across both academia and industry. Yet, despite the remarkable progress which has been made towards generation techniques, the evaluation of the fidelity of the synthetic data remains a critical open challenge [7]. More specifically, the existing synthetic data fidelity assessment pipelines [13, 14] are often fragmented, relying on ad-hoc combinations of summary statistics, simple distributional tests, or domain-specific heuristics. Such approaches rarely capture the deeper multivariate, nonlinear, or topological characteristics of data which are essential for high utility and for ensuring privacy-preservation. Furthermore, early synthetic data evaluation methods typically rely on the comparison of empirical means, variances, and univariate distributions using tests such as the Kolmogorov–Smirnov or $\chi^2$ statistic [14]. While these methods provide a sufficient and preliminary insight into the data structure, they are inherently limited, as they cannot characterize high-dimensional dependencies, non-linear relationships, multimodal behavior, or structural patterns. Metrics such as the Jensen–Shannon divergence, the Wasserstein distance, the Hellinger distance, and the mutual information differences [14] represent more advanced alternatives; yet, in practice, most evaluations remain feature-centric and focus only on isolated distributions rather than capturing the holistic structure of the dataset. Similarly, multivariate statistics, including covariance similarity, correlation matrix distance, and rank-based correlations are frequently used to assess the preservation of dependence. However, these approaches typically examine linear or pairwise interactions, therefore overlooking complex higher-order relationships, heterogeneous data types, and mixed categorical–numerical structures that exist in real-world data.

To address these limitations, recent research studies have explored embedding-based evaluation [15], where the tabular data are projected into learned latent spaces via autoencoders, contrastive models, transformers, or similar representation-learning methods. The similarity is then quantified using cosine distance, nearest-neighbor overlap, or measures such as Centered Kernel Alignment (CKA) [16]. These methods offer richer geometric insights by reflecting global manifold properties. However, they introduce new challenges which are related to model dependence, computational cost, and interpretability, particularly for mixed-type tabular data, where standardized encoders are less mature. Graph-based structural fidelity is another option, where k-nearest neighbor graphs, correlation networks, or mutual information graphs are used to capture the dataset's topology [14]. More specifically, comparisons based on graph Laplacian spectra, neighborhood overlaps, or structural distances can detect subtle distortions such as mode collapse, oversmoothing, or loss of local geometric structure [14]. Despite the progress in synthetic data fidelity assessment, several recurring limitations persist. The current approaches remain fragmented, where the statistical, embedding-based, and graph-theoretic metrics are being typically assessed "in isolation" rather than through a unified methodology. The evaluation criteria also vary widely across the studies, which results in inconsistent metric selection that complicates cross-model or cross-dataset comparison. Moreover, the existing toolkits provide insufficient support for mixed-type tabular data, particularly when handling categorical features, multi-level categories, and their associated dependency structures. In addition, graph-theoretic metrics remain underutilized and are rarely integrated into existing toolkits [13, 17, 18]. Data visualization capabilities are often limited, offering quantitative outputs but lacking interpretable diagnostics to help practitioners understand where and why synthetic data deviates from real data. These gaps obscure the reliable use of synthetic data in critical sectors, where transparency and auditability are critical.

To address these gaps, we introduce the Synthetic Data Blueprint (SDB) framework [19], implemented as a modular and extensible Pythonic-based library to support a "multi-view" evaluation of the fidelity of synthetic tabular data. SDB unifies univariate and multivariate statistical measures, categorical association metrics, embedding-based similarity analysis, graph topology preservation indicators, and privacy-oriented distance assessments into a single, coherent manner. The SDB framework is built on statistical and computational principles and aims to foster transparency and auditability through an extensive suite of visual diagnostics and structured JSON reporting. It ensures robustness across mixed data types, including numerical, binary, and multi-categorical features. In addition, it provides standardized and repeatable pipelines that allow consistent benchmarking. It also contributes to the maturity, safety, and trustworthiness of synthetic data, to support research, industrial deployment, regulatory auditing, and alignment, which is in line with emerging trustworthy AI frameworks like the AI-Act. To demonstrate the robust and domain-agnostic applicability of the SDB, we applied it across three representative real-world use cases, including: (i) healthcare, (ii) socioeconomic and financial modelling, and (iii) cybersecurity network analysis. The empirical findings across the three use cases reveal that the SDB not only captures distributional alignment but also exposes deeper structural behaviors of synthetic data generators, such as sensitivity to skewed clinical variables, robustness under high-cardinality categorical domains, and resilience to heavy-tailed telemetry signals. Moreover, our findings demonstrate how the SDB can help practitioners to understand why deviations occur in synthetic data and to assess whether these are acceptable or not in real-world applications.

## 2. The Synthetic Data Blueprint (SDB) framework

The SDB is designed as a unified, modular, and extensible framework for assessing the fidelity, structure, and privacy characteristics of synthetic tabular data. It integrates statistical, topological, and embedding-based measures with metadata quality assessments and a thorough visualization suite. It aims to standardize synthetic data evaluation by providing a reproducible pipeline that captures both surface-level distributional properties and deeper structural relationships within the data. This section provides an overview of SDB's design principles, core components, and evaluation workflow.

### 2.1. Design principles

SDB is built upon four foundational principles: (i) mathematical formulations, to ensure that all metrics are expressed in a proper mathematical way for reproducibility, (ii) transparency, where structured JSON reports and visualization tools are used to help practitioners interpret the fidelity assessment results, (iii) mixed-type robustness, for the consistent manipulation of numerical, binary, ordinal, and multi-categorical variables, and (iv) modularity and extensibility, through the integration of new metrics, plots, or privacy risk indicators as synthetic data science evolves. These principles make SDB suitable for both research and production contexts, including regulated environments that require reproducible and auditable evaluation pipelines.

### 2.2. Architecture

The SDB accepts as input a configuration file (in. yaml format), where the user defines the following mandatory parameters: (i) the path to the real tabular dataset in .csv format (with dimensions $MxN$, where $M$ denotes the number of rows and $N$ the number of columns), (ii) the path to the synthetic dataset (in the same format), (iii) the name of the output report (in. JSON), and (iv) the folder where to store the plots. The rest of the (optional) parameters appear in the Appendix. SDB adopts a modular architecture which allows its' components to operate independently while also contributing to the generation of a unified assessment report. According to Fig. 1, the architecture of the SDB framework is organized into four primary modules, namely: (i) Module 1 - Data quality assessment (DQA), which focuses on the automated identification of feature (or variable) types, missing values, and outliers within the real input dataset, (ii) Module 2 - Statistical fidelity evaluation (SFE), which aims to compare the statistical properties between the real and synthetic datasets using a diverse set of distribution-level and dependency metrics, (iii) Module 3 - Structural and topological assessment (STA), which aims to leverage embeddings and graph-based representations to capture manifold-level differences between the real and synthetic datasets, (iv) Module 4 - Reporting and visualization, where all the data fidelity assessment results are aggregated into human-readable plots and machine-readable JSON outputs.

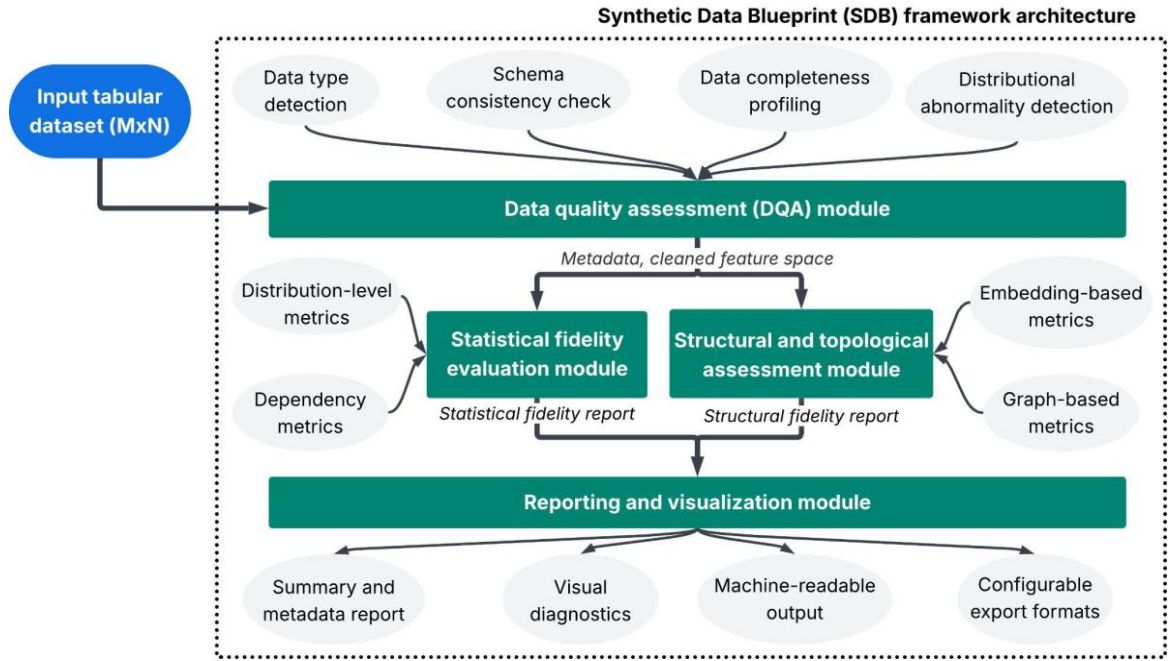

**Figure 1.** The SDB framework architecture.

## 2.2. Data quality assessment (DQA) module

The Data Quality Assessment (DQA) module is a core aspect in the SDB. Its purpose is to perform a systematic examination of the input tabular dataset by: (i) characterizing its structure, (ii) identifying potential quality issues, and (iii) preparing a standardized feature space for fidelity assessment. The module begins by performing data type detection, to automatically classify features as numerical, binary categorical, multi-categorical, or textual. This ensures that statistical and structural metrics are applied appropriately and consistently across feature types. In parallel, the DQA module conducts a schema consistency check, to verify that real and synthetic data share a harmonized structure, including aligned feature names, compatible datatypes, and expected value domains. Any discrepancies are recorded, and only standard-aligned features are propagated for evaluation. In addition to structural validation, the module produces a detailed data completeness profile, which quantifies missing values at both global and per-feature levels. This assessment helps to contextualize fidelity outcomes by revealing portions of the dataset that may inherently limit reproducibility. The DQA module further performs distributional abnormality detection, relying on robust statistical methods (e.g., IQR-based outlier detection) to identify anomalous or irregular patterns, such as, extreme values, unexpected sparsity, or category imbalance. These are crucial to determine whether deviations between real and synthetic data originate from generator limitations or from imperfections in the quality of real-world data. All these processes yield a thorough metadata block that summarizes dataset characteristics, including feature distributions, missingness, and outlier statistics, and generate a clean and harmonized feature space ready for analysis. This output serves as the structured input to the Statistical Fidelity Evaluation module and the Structural & Topological Assessment module, which are presented next.

## 2.3. Statistical fidelity evaluation (SFE) module

The statistical fidelity evaluation (SEF) module aims to quantify how well the synthetic data replicate the statistical properties of the real data. It operates on the cleaned feature space which is delivered by the DQA module, and it is organized into two complementary components (i.e. collections of fidelity assessment metrics): (i) the distribution-level metrics, and (ii) the dependency metrics. These two components evaluate not only the fidelity of individual feature distributions but also the preservation of inter-feature relationships that define the dataset's structural integrity. They are both used to generate the "Statistical fidelity report", which captures per-feature divergences, multivariate structural alignment measures, and global metrics summarizing the similarity between the real and synthetic data.

### 2.3.1. Distribution-level metrics

These metrics aim to evaluate the similarity between the real and synthetic data at the feature (univariate) level. For every common feature in the aligned datasets, the module computes a diverse set of statistical divergence measures. For the numerical features, the module assesses distributional fidelity through: (i) the Kolmogorov–Smirnov (KS) statistic, which captures the maximum deviation between empirical CDFs, (ii) the Kullback-Leibler divergence (KLD), which quantifies how much information is lost when approximating the real distribution with the synthetic one, (iii) the Jensen–Shannon divergence (JSD), which is a symmetric, smoothed divergence functional for noisy distributions, (iv) the Wasserstein distance (WD), which quantifies the minimal effort required to morph one distribution into another, (v) the Hellinger distance (HD), which is sensitive to differences in distribution shape, and (vi) the total variation distance (TVD), which captures the maximum discrepancy in probability mass. For categorical features, the module computes: (i) the Range Coverage (RC), which evaluates whether the synthetic dataset reproduces the full value span of the real dataset by measuring the degree of overlap between their numerical ranges, (ii) the Chi-square statistic (CSS), which aims to test the consistency of category frequencies, (iii) the Category coverage (CC), which measures whether all real categories appear in the synthetic dataset, and (iv) the Cramér's V (CV), which compares the categorical distribution strength. Through these metrics we ensure that both continuous and discrete statistical properties are faithfully analyzed. The output for each feature is stored in the "local_metrics" field of the JSON report.

#### 2.3.1.1. General formulation

Let $X = \{x_1, ..., x_n\}$ denote the real samples of a feature, $Y = \{y_1, ..., y_m\}$ denote the synthetic samples of the same feature. Let $F_X(t)$ and $F_Y(t)$ denote the empirical cumulative distribution functions (ECDFs) of $X$ and $Y$, respectively, as in:

$$F_X(t) = \frac{1}{n} \sum_{j=1}^{n} \mathbf{1}\{x_j \leq t\}, F_Y(t) = \frac{1}{m} \sum_{j=1}^{m} \mathbf{1}\{y_j \leq t\}. \qquad (1)$$

For categorical features with categories $\{c_1, ..., c_K\}$, let $P = (p_1, ..., p_K)$ and $Q = (q_1, ..., q_K)$ denote the empirical probability mass functions (PMFs), where $p_i = Pr(X = c_i)$ and $q_i = Pr(Y = c_i)$.

#### 2.3.1.2. Kolmogorov–Smirnov (KS) statistic

To measure the maximum discrepancy between the ECDFs of the real and synthetic data, the Kolmogorov–Smirnov (KS) statistic is defined as in:

$$KS(X, Y) = \sup_{t \in \mathbb{R}} |F_X(t) - F_Y(t)|, \qquad (2)$$

where $F_X(t)$ and $F_Y(t)$ are the ECDFs of the real and synthetic samples. The supremum $\sup_t$ seeks the point where the absolute difference between the ECDFs is largest to capture the strongest deviation.

#### 2.3.1.3. Kullback-Leibler divergence (KLD)

The Kullback–Leibler Divergence (KLD) of the empirical probability mass function (PMF) of the real dataset, $P = (p_1, ..., p_K)$ from the empirical PMF of the synthetic dataset $Q = (q_1, ..., q_K)$, say $KL(P \parallel Q)$, is defined as in:

$$KL(P \parallel Q) = \sum_{i=1}^{K} p_i \, log\left(\frac{p_i}{q_i}\right), \tag{3}$$

where $p_i$ is the probability that the real dataset takes value in bin or category $i$, and $q_i$ is the probability that the synthetic dataset takes value in bin or category $i$. A value $KL(P \parallel Q) = 0$ if and only if $P = Q$. A value $KL(P \parallel Q) > 0$ indicates divergence between real and synthetic distributions. The metric is asymmetric meaning that $KL(P \parallel Q) \neq KL(Q \parallel P)$ which reflects directional information loss.

*2.3.1.4. Jensen–Shannon divergence (JSD)*

To quantify a symmetric, smoothed divergence between the real and synthetic probability distributions, the mixture distribution is first defined as in:

$$M = \frac{1}{2}(P + Q) = (m_1, \ldots, m_K), m_i = \frac{1}{2}(p_i + q_i), \tag{4}$$

where $p_i$ and $q_i$ are the empirical probabilities for each category/bin, and $M$ is the midpoint ("average distribution"). The Jensen–Shannon divergence (JSD) is then defined as in:

$$JS(P, Q) = \frac{1}{2} KL(P \parallel M) + \frac{1}{2} KL(Q \parallel M), \tag{5}$$

where $KL(P \parallel M)$ is the Kullback-Leibler divergence between $P$ and $M$. JSD is finite and symmetric, capturing distributional dissimilarity even for noisy or multimodal distributions.

*2.3.1.5. Wasserstein distance (WD)*

To measure the minimum mass-transport cost needed to transform one empirical distribution into another, the 1-Wasserstein distance (WD), $W_1(X, Y)$, is defined as in:

$$W_1(X, Y) = \int_{-\infty}^{\infty} |F_X(t) - F_Y(t)| \, dt. \tag{6}$$

For empirical data with sorted samples $x_{(1)} \leq \cdots \leq x_{(n)}$ and $y_{(1)} \leq \cdots \leq y_{(n)}$ (assuming $n = m$ or using interpolation) the 1-Wasserstein distance is defined as in:

$$W_1(X, Y) = \frac{1}{n} \sum_{i=1}^{n} |x_{(i)} - y_{(i)}|, \tag{7}$$

where $x_{(i)}$ and $y_{(i)}$ represent quantile-matched samples, making the metric sensitive to shifts in distribution shape and support.

*2.3.1.6. Hellinger distance (HD)*

To quantify the geometric divergence between probability distributions using square-root densities, the Hellinger distance (HD) is utilized which is defined as in:

$$H(P, Q) = \frac{1}{\sqrt{2}} \left( \sum_{i=1}^{K} (\sqrt{p_i} - \sqrt{q_i})^2 \right)^{\frac{1}{2}}, \tag{8}$$

where $\sqrt{p_i}$ and $\sqrt{q_i}$ form the vectors in the Hellinger geometry. The distance is bounded in the range [0,1], where 0 indicates identical distributions.

*2.3.1.7. Total Variation Distance (TVD)*

To measure the maximum possible difference in assigned probability mass between two distributions, the total variation distance (TVD) is defined as in:

$$TVD(P, Q) = \frac{1}{2} \sum_{i=1}^{K} |p_i - q_i|, \tag{9}$$

where TVD corresponds to the largest difference in probabilities assigned to any measurable set. The factor 1/2 ensures that the TVD lies in [0,1].

*2.3.1.8. Range Coverage (RC)*

To evaluate whether the synthetic dataset reproduces the full value range of the real dataset for a numerical feature, we define the Range Coverage (RC) as the ratio between the length of the intersection of real and synthetic value ranges and the length of the real value range. Let $range_X = [min(X), max(X)]$ be the real-data range, and $range_Y = [min(Y), max(Y)]$ be the synthetic-data range. The Range Coverage (RC) is defined as in:

$$RC(X,Y) = \frac{max(0, min(max(X), max(Y)) - max(min(X), min(Y)))}{max(X) - min(X)}, \quad (10)$$

where $min(X), max(X)$ denote the minimum and maximum values of the feature in the real dataset, $min(Y), max(Y)$ denote the minimum and maximum values in the synthetic dataset. The numerator computes the length of the overlapping interval between the two ranges. If the ranges do not overlap, the numerator becomes 0. The denominator normalizes by the length of the real-data range, ensuring that $RC(X,Y) \in [0,1]$. A value $RC = 1$ denotes that the synthetic dataset fully covers the entire real-data range. A value $RC = 0$ denotes that the synthetic dataset's range does not overlap with the real range at all. Values $0 < RC < 1$ denote partial coverage (missing extremes or truncated distributions).

*2.3.1.9. Chi-square statistic (CSS)*

To test whether the synthetic category frequencies match the real ones, the Chi-square statistic, $\chi^2$, is defined as in:

$$\chi^2 = \sum_{i=1}^{K} \frac{(O_i - E_i)^2}{E_i}, \quad (11)$$

where $O_i$ and $E_i$ denote synthetic (observed) and real (expected) counts for category $c_i$. Large values indicate stronger deviation between synthetic and real category frequencies.

*2.3.1.10. Category Coverage (CC)*

To quantify whether the synthetic data contain all the categories present in the real data, we define the Category Coverage (CC) as in:

$$CC(X,Y) = \frac{|\mathcal{C}_{real} \cap \mathcal{C}_{synthetic}|}{|\mathcal{C}_{real}|}, \quad (12)$$

where $\mathcal{C}_{real}$ and $\mathcal{C}_{synthetic}$ denote the sets of categories appearing in real and synthetic data. A value $CC = 1$ means that every real category appears at least once in the synthetic dataset. It is useful for the detection of mode dropouts or generator sparsity.

*2.3.1.11. Cramér's V (CV)*

To quantify the magnitude of discrepancy between real and synthetic categorical distributions, the Cramér's $V$ (CV) is computed from $\chi^2$ as in:

$$CV = \sqrt{\frac{\chi^2}{n(K-1)}}, \quad (13)$$

where $n$ is the sample size of the real dataset, and $K$ is the number of distinct categories. A lower $CV$ indicates closer alignment between real and synthetic category distributions (in the univariate case).

*2.3.2. Dependency metrics*

These metrics aim to evaluate multivariate structural fidelity by assessing whether the relationships between variables are preserved in the synthetic dataset; a crucial property for modeling, inference, and causal analysis. For numerical dependencies, the module computes: (i) the covariance matrix similarity (CMS) or Frobenius norm difference, which indicates overall shape preservation in the multivariate distribution, (ii) the correlation matrix distance (CMD), which quantifies the global alignment between real and synthetic correlation structures, (iii) the Correlation difference (both the Pearson – CDP, and the Spearman - CDS) per pair of variables to assess linear and monotonic relationship consistency. For categorical and mixed-type dependencies, the module computes the Mutual Information Difference (MID), which evaluates whether nonlinear associations are maintained across datasets. These metrics collectively ensure that higher-order interactions, beyond marginal distributions, remain intact. To this end, the SDB detects subtle distortions such as broken correlations, inflated associations, or missing variable interactions that could render synthetic data unreliable for downstream analytical tasks.

*2.3.2.1. General formulation*

Let the real dataset be $X = (X_1, X_2, \ldots, X_d)$ and the synthetic dataset be $Y = (Y_1, Y_2, \ldots, Y_d)$, with $d$ features in total. Let $\Sigma_X$ and $\Sigma_Y$ be the empirical covariance matrices of the real and synthetic numerical features, $R_X$ and $R_Y$ be the corresponding Pearson correlation matrices, $\rho_{ij}^{(P)}$ and $\rho_{ij}^{(S)}$ be the Pearson and Spearman correlations between features $i$ and $j$, $I_X(i,j)$ and $I_Y(i,j)$ denote the empirical mutual information between features $i$ and $j$ in real and synthetic data, $P_{ij}$ and $Q_{ij}$ be joint categorical distributions for a pair of discrete features.

*2.3.2.2. Covariance Matrix Similarity (Frobenius Norm Distance) (CMS)*

To measure how closely the synthetic dataset preserves the multivariate shape of the real dataset, the Frobenius distance between the covariance matrices is defined as in:

$$D_{\text{cov}} = \| \Sigma_X - \Sigma_Y \|_F = \left( \sum_{i=1}^{d} \sum_{j=1}^{d} (\Sigma_{X,ij} - \Sigma_{Y,ij})^2 \right)^{1/2}, \qquad (14)$$

where, $\Sigma_X, \Sigma_Y$ are the covariance matrices of real and synthetic data, $\Sigma_{X,ij}$ is the covariance between features $i$ and $j$ in the real dataset, and $\|\cdot\|_F$ is the Frobenius matrix norm. Lower values imply better preservation of multivariate variability and scale structure.

*2.3.2.3. Correlation Matrix Distance (CMD)*

To quantify the global alignment in the pairwise linear associations, the correlation matrix distance (CMD) is computed as in:

$$CMD = \frac{\| R_X - R_Y \|_F}{\| R_X \|_F}, \qquad (15)$$

where $R_X, R_Y$ are the Pearson correlation matrices of the real and synthetic datasets, and $\|\cdot\|_F$ is the Frobenius matrix norm. Normalization ensures scale-free comparison. Lower values imply similar correlation structures.

*2.3.2.4. Correlation Difference (Pearson) (CDP)*

To assess how well the synthetic data preserve the linear and monotonic relationships between individual feature pairs, we utilize the Pearson correlation difference which is defined as in:

$$CDP_{ij}^{(P)} = |\rho_{X,ij}^{(P)} - \rho_{Y,ij}^{(P)}|, \tag{16}$$

where $\rho_{X,ij}^{(P)}$ is the Pearson correlation between features $i$ and $j$ in the real data, and $\rho_{Y,ij}^{(P)}$ is the Pearson correlation between features $i$ and $j$ in the synthetic data.

*2.3.2.5. Correlation Difference (Spearman) (CDS)*

The Spearman correlation difference (CDS) is defined as in:

$$CDS_{ij}^{(S)} = |\rho_{X,ij}^{(S)} - \rho_{Y,ij}^{(S)}|, \tag{17}$$

where $\rho_{X,ij}^{(S)}$ is the Spearman rank correlation (monotonic association) between features $i$ and $j$ in the real data, and $\rho_{Y,ij}^{(P)}$ is the Spearman rank correlation between features $i$ and $j$ in the synthetic data.

*2.3.2.6. Mutual Information Difference (MID)*

To evaluate whether the non-linear associations between the features in the real and the synthetic data are preserved, the mutual information difference is defined as:

$$MID_{ij} = |I_X(i,j) - I_Y(i,j)|, \tag{18}$$

where $I_X(i,j)$, $I_Y(i,j)$ is the mutual information between features $i$ and $j$ in $X$ and $Y$, respectively. The MID captures non-linear, non-monotonic dependencies. Large differences indicate broken or spurious associations.

### 2.4. Structural and topological assessment (STA) module

The Structural and Topological Assessment (STA) Module aims to evaluate synthetic data fidelity, by examining not only the statistical properties of individual features but also the geometric, structural, and topological organization of the dataset. While statistical metrics quantify similarity at the distribution and dependency levels, this module captures how samples are arranged in the underlying data manifold; a critical aspect for generative models, clustering behavior, and machine learning tasks. To achieve this, the module transforms the tabular dataset into a set of latent embeddings and then computes structural fidelity using both embedding-based and graph-based metrics. Both jointly quantify the similarity between the real and synthetic datasets in terms of learned representations, local neighborhoods, and global manifold structure. The STA module generates a "Structural fidelity report", which consists of: (i) embedding-level similarity scores, (ii) graph-level structural comparisons, (iii) a unified EGFS score, and (iv) visual diagnostics (PCA overlays, UMAP/TSNE embeddings, kNN graph comparisons). These results are processed by the Reporting and Visualization module and integrated into both human-readable and machine-readable outputs.

2.4.1. Embedding-based metrics

This set of metrics evaluates how closely the synthetic dataset replicates the structure of the real dataset in the embedding space. The module constructs embeddings, which perform: (i) Standardization and PCA compression for numerical features, (ii) Frequency encoding + PCA for categorical features, (iii) TF-IDF + SVD compression for text fields (when present), and (iv) L2 normalization, ensuring compatibility with cosine similarity–based structural metrics. These embeddings provide a compact latent representation of the dataset, enabling the comparison of global geometric properties. The following embedding-level metrics are computed: (i) Centered Kernel Alignment (CKA), which quantifies the similarity between the real and synthetic embedding matrices, (ii) Average Wasserstein Embedding Distance (AWED), which computes the 1-D Wasserstein distance for each embedding dimension and averages across all dimensions. Together, these metrics provide a direct measure of how well the synthetic dataset matches the *latent structure* of the real dataset.

After the preprocessing pipeline which includes standardization, PCA compression for numeric features, frequency encoding + PCA for categorical features, TF–IDF + SVD for text fields (when applicable), and final $L^2$-normalization; the real and synthetic datasets are mapped into latent embedding matrices:

$$Z_X \in \mathbb{R}^{n \times k}, Z_Y \in \mathbb{R}^{m \times k}, \tag{19}$$

where $n, m$ are the number of real and synthetic samples, $k$ is the embedding dimensionality, rows of $Z_X, Z_Y$ are normalized so that $\| z_{i,\cdot} \|_2 = 1$. These embeddings represent each sample in a common latent space, enabling geometric comparisons between real and synthetic data.

*2.4.1.1. Centered Kernel Alignment (CKA)*

To quantify the similarity of global representation geometry between real and synthetic embeddings, the linear CKA similarity is first computed as in:

$$CKA(Z_X, Z_Y) = \frac{\| Z_X^\top Z_Y \|_F^2}{\| Z_X^\top Z_X \|_F \, \| Z_Y^\top Z_Y \|_F}, \tag{20}$$

where $Z_X, Z_Y$ are embedding matrices constructed from real and synthetic datasets, $Z_X^\top Z_Y$ is the cross-covariance of embeddings, and $\|\cdot\|_F$ is the Frobenius norm. The numerator measures alignment strength between embedding spaces. The denominator normalizes for scale and orthogonal transformations. $CKA \in [0,1]$, where values near 1 indicate that real and synthetic datasets share the same global representation geometry, independent of rotation or rescaling.

*2.4.1.2. Average Wasserstein Embedding Distance (AWED)*

To evaluate how well the distribution of embedded points is preserved in each latent dimension, we compute the 1-dimensional Wasserstein distance for each embedding coordinate. Let $Z_X^{(j)} = (Z_{X,1j}, \dots, Z_{X,nj}), Z_Y^{(j)} = (Z_{Y,1j}, \dots, Z_{Y,mj})$, denote the $j$-th embedding dimension of real and synthetic datasets, for $j = 1, \dots, k$. The embedding-level Wasserstein distance (WED) is defined as in:

$$WED_j = W_1(Z_X^{(j)}, Z_Y^{(j)}) = \int_{-\infty}^{\infty} | F_X^{(j)}(t) - F_Y^{(j)}(t) | \, dt, \tag{21}$$

where $F_X^{(j)}$ and $F_Y^{(j)}$ are the ECDFs of the $j$-th embedding dimension. The Average Wasserstein Embedding Distance (WED) is then defined as:

$$AWED = \frac{1}{k}\sum_{j=1}^{k} W_j, \qquad (22)$$

where $Z_X^{(j)}$, $Z_Y^{(j)}$ are the $j$-th coordinate of the embedded real and synthetic datasets, $W_j$ is the Wasserstein distance in latent dimension $j$, $F_X^{(j)}(t)$ ECDF of the $j$-th embedding coordinate. The averaging across all embedding dimensions yields the final global distributional similarity score in the latent space. Lower AWED values indicate that synthetic embeddings closely match real embeddings across all latent dimensions.

2.4.2. Graph-based metrics

The second set of metrics evaluates the topological similarity between the real and synthetic datasets by analyzing graph structures derived from their latent embeddings. After constructing a $k$-nearest neighbor (kNN) graph for each dataset, thus forming a topological approximation of the underlying data manifold, the module quantifies fidelity across both the local neighborhood structure and the global graph geometry. Specifically, it computes: (i) the Neighborhood Overlap (Jaccard Similarity), which measures the proportion of shared neighbors between the real and synthetic graphs, (ii) the Spectral Distance, which compares the eigenvalue spectra of the normalized graph Laplacians to assess global topological alignment, and (iii) the Graph Structural Fidelity Score (GSFS), which evaluates similarity in degree, clustering, and path-length distributions. These metrics capture how faithfully the synthetic dataset preserves both the local manifold geometry and the global structural organization of the real data in embedding space.

Let the real and synthetic datasets be embedded into latent matrices $Z_X \in \mathbb{R}^{n \times k}, Z_Y \in \mathbb{R}^{m \times k}$, as defined in Section 2.4.1. For each dataset, a k-nearest neighbor graph (kNN graph) is constructed: $G_X = (V_X, E_X)$ from $Z_X$, $G_Y = (V_Y, E_Y)$ from $Z_Y$, where nodes correspond to samples and edges connect each sample to its $k$ nearest neighbors in embedding space. Let: $N_X(i)$ be the set of $k$-nearest neighbors of node $i$ in the real kNN graph, and $N_Y(i)$ be the corresponding set in the synthetic kNN graph.

2.4.2.1. Neighborhood Overlap (NO) (Jaccard Similarity)

To measure how well local manifold structure is preserved, we compute the Jaccard similarity between neighborhoods of corresponding points:

$$J(i) = \frac{|N_X(i) \cap N_Y(i)|}{|N_X(i) \cup N_Y(i)|}, \qquad (24)$$

where $N_X(i)$ are the indices of the $k$-nearest neighbors of sample $i$ (real data), $N_Y(i)$ are the indices of the $k$ nearest neighbors of corresponding synthetic sample. The Neighborhood Overlap (NO) score is defined as the mean Jaccard index across all matched nodes:

$$NO = \frac{1}{n}\sum_{i=1}^{n} J(i), \qquad (25)$$

where $J(i)$ is the Jaccard similarity for node $i$. A value $NO = 1$ denotes perfect preservation of local neighborhoods, whereas a value $NO = 0$ denotes no shared neighbors. This metric evaluates local topological fidelity in the embedding manifold.

2.4.2.2. Spectral Distance (SD) Between Graph Laplacians

To compare global topology, including cluster connectivity and manifold structure, we compute the difference between the eigenvalue spectra of the normalized graph Laplacians. The adjacency matrices are first constructed as in:

$$A_X, A_Y \in \mathbb{R}^{n \times n}.$$

Let $D_X = diag(d_{X,1}, \ldots, d_{X,n})$ with $d_{X,i} = \sum_j A_{X,ij}$, $D_Y$ defined analogously.

The normalized Laplacians are then defined as in:

$$L_X = I - D_X^{-1/2} A_X D_X^{-1/2}, L_Y = I - D_Y^{-1/2} A_Y D_Y^{-1/2}. \tag{26}$$

Let $\lambda_X = (\lambda_{X,1}, \ldots, \lambda_{X,n})$ and $\lambda_Y = (\lambda_{Y,1}, \ldots, \lambda_{Y,n})$ be the eigenvalues of $L_X$ and $L_Y$, sorted in ascending order. The Spectral Distance (SD) is defined as in:

$$SD = \| \lambda_X - \lambda_Y \|_2 = \left( \sum_{i=1}^{n} (\lambda_{X,i} - \lambda_{Y,i})^2 \right)^{1/2}, \tag{27}$$

where $A_X, A_Y$ are the adjacency matrices of the real and synthetic kNN graphs, $L_X, L_Y$ are the normalized Laplacians which capture the global graph structure, $\lambda_{X,i}$ are the eigenvalues encoding connectivity, cluster separation, and diffusion structure. This metric evaluates global topological fidelity, where low $SD$ values denote that the synthetic data preserve the global manifold structure whereas high $SD$ values denote topological distortions.

*2.4.2.3. Graph Structural Fidelity Score (GSFS)*

To measure how well global structural properties of the real kNN graph are preserved, we compare key graph-theoretic statistics. Let $d_X, d_Y$ denote the degree distributions, $c_X, c_Y$ denote the clustering coefficient distributions, $s_X, s_Y$ denote the shortest-path length distributions. Three similarity components $S_d$, $S_c$, and $S_s$ are defined using normalized Frobenius or $\ell_2$ distances, where $S_d$ is the similarity of degree distributions (connectivity strength), $S_c$ is the similarity of local clustering (triadic density), and $S_s$ is the similarity in global connectivity via shortest paths:

$$S_d = 1 - \frac{\| d_X - d_Y \|_2}{\| d_X \|_2}, S_c = 1 - \frac{\| c_X - c_Y \|_2}{\| c_X \|_2}, S_s = 1 - \frac{\| s_X - s_Y \|_2}{\| s_X \|_2}. \tag{28}$$

The Graph Structural Fidelity Score (GSFS) is the weighted average of the similarity components:

$$GSFS = \alpha_d S_d + \alpha_c S_c + \alpha_s S_s, \tag{29}$$

with default weights:

$$\alpha_d = \alpha_c = \alpha_s = \frac{1}{3}. \tag{30}$$

The GSFS values lie in [0,1], where higher values indicate better preservation of global topology. GSFS captures manifold-level structure, complementing the NO (local) and the SD (spectral/global).

## 2.5. Summary of metrics

A summary of the supported metrics from Sections 2.3-2.4 is presented in Table 1. The SDB framework integrates a diverse suite of fidelity metrics from distribution-level, dependency-level, to embedding-based, and graph-based. Distributional similarity is quantified through classical divergence measures, including KS, JSD, KLD, WD, HD, and TVD, alongside categorical-specific indicators such as CSS, CC, and RC. Dependency preservation is assessed using CDP and CDS, MID, CMS, and CV for categorical associations. To capture deeper structural and topological alignment, the SDB incorporates embedding-based metrics such as CKA and AWED, as well as, graph-theoretic measures including the NO, the SD, and the GFS.

**Table 1.** A summary of the SDB-supported metrics for fidelity assessment.

| No | Metric name | Acronym | Type of metric | Type of supported data | Description |
|---|---|---|---|---|---|
| 1 | Kolmogorov–Smirnov Statistic | KS | Distribution-Level Metrics | Continuous | Measures the maximum distance between the empirical cumulative distributions of real and synthetic data for a numeric feature. |
| 2 | Kullback-Leibler Divergence | KLD | Distribution-Level Metrics | Continuous / Categorical | Quantifies how much information is lost when approximating the real data distribution with the synthetic one. Asymmetric measure. |
| 3 | Jensen–Shannon Divergence | JSD | Distribution-Level Metrics | Continuous / Categorical | Symmetric measure of similarity between two probability distributions derived from real and synthetic data. Lower values indicate higher similarity. |
| 4 | Wasserstein Distance (Earth Mover's) | WD | Distribution-Level Metrics | Continuous | Quantifies the minimum "work" required to transform one probability distribution into another, reflecting both shape and distance differences. |
| 5 | Hellinger Distance | HD | Distribution-Level Metrics | Continuous / Categorical | Measures the distance between two probability distributions; bounded between 0 (identical) and 1 (completely dissimilar). |
| 6 | Total Variation Distance | TVD | Distribution-Level Metrics | Continuous / Categorical | The Total Variation Distance measures the maximum difference between two probability distributions. |
| 7 | Range Coverage | RC | Distribution-Level Metrics | Continuous | The Range Coverage metric measures how much of the real data's value range is covered by the synthetic data. |
| 8 | Chi-Square Statistic | CSS | Distribution-Level Metrics | Categorical | Tests whether the observed category frequencies in the synthetic data differ significantly from those in the real data. |
| 9 | Category coverage | CC | Distribution-Level Metrics | Categorical | Proportion of unique categories in the real data that also appear in the synthetic data; detects missing or underrepresented categories. |
| 10 | Contingency Table Similarity (Cramér's V) | CV | Dependency Metrics | Categorical | Measures the strength of association between two categorical variables in real vs. synthetic datasets; used to compare inter-feature dependencies. |
| 11 | Covariance Matrix Similarity (Frobenius Norm) | CMS | Dependency Metrics | Continuous | Quantifies deviation between real and synthetic covariance matrices; smaller Frobenius norm indicates closer similarity. |
| 12 | Correlation Matrix Distance | CMD | Dependency Metrics | Continuous | Computes normalized Frobenius norm of the difference between correlation matrices; used as an overall measure of structural fidelity. |
| 13 | Correlation Difference (Pearson) | CDP | Dependency Metrics | Continuous / Categorical | Measures how much the linear correlations between features differ between real and synthetic datasets. |
| 14 | Correlation Difference (Spearman) | CDS | Dependency Metrics | Continuous / Categorical | Measures how much the rank correlations between features differ between real and synthetic datasets. |
| 15 | Mutual Information Difference | MID | Dependency Metrics | Continuous / Categorical | Captures how well nonlinear dependencies between variables are preserved; compares mutual information matrices between real and synthetic data. |
| 16 | Centered Kernel Alignment | CKA | Embedding-Based Metrics | Continuous / Categorical | Kernel-based similarity metric comparing representation matrices (embeddings) of real and synthetic data. Robust to isotropic scaling; higher CKA indicates that real and synthetic datasets encode similar feature relationships. |
| 17 | Average Wasserstein Embedding Distance | AWED | Embedding-Based Metrics | Continuous / Categorical | Computes the average Wasserstein distance between embedded representations (e.g., UMAP, PCA, t-SNE) of real and synthetic samples. Lower values indicate that the synthetic embedding closely matches the real one across global geometry. |
| 18 | Neighbor Overlap (Jaccard Similarity) | NO | Graph-Based Metrics | Continuous / Categorical | Measures the similarity between nearest-neighbor sets of real and synthetic samples (e.g., k-NN neighborhoods). Computed as the Jaccard index: intersection over union of neighbor sets. Higher overlap indicates better local structure preservation. |
| 19 | Spectral Distance | SD | Graph-Based Metrics | Continuous / Categorical | Compares the spectra (eigenvalues) of graph Laplacians derived from real and synthetic data. Captures global geometric and manifold differences. Smaller spectral distance means the |

| | | | | synthetic data preserves the intrinsic structure of the real data. |
|---|---|---|---|---|
| 20 | Graph Structural Fidelity Score | GSFS | Graph-Based Metrics | Continuous / Categorical | Assesses whether the structural properties of real and synthetic data graphs (e.g., k-NN graph, similarity graph) are preserved—including degree distribution, clustering coefficients, and connectivity patterns. Higher scores indicate better structural fidelity. |

## 2.6. Reporting and visualization module

The Reporting and Visualization Module serves as the final layer of the framework, responsible for transforming the raw analytical outputs generated across all previous modules into structured reports, visual summaries, and interpretable artifacts. Its purpose is to ensure that the results of the fidelity assessment are not only machine-readable for automated processes but also human-interpretable for researchers, data scientists, and domain experts. The module integrates numerical metrics, graphical diagnostics, metadata, and textual explanations into an accessible format, to support informed decision-making about the quality and reliability of synthetic datasets. At the core of this module is the generation of a run-specific JSON report, to ensure full traceability and reproducibility of assessments. The module automatically assigns each evaluation a unique "run_id", and all the results, including distribution-level, dependency, structural, and topological metrics, as well as dataset metadata, are serialized into a JSON file in the leading execution directory. This report includes a diverse set of metric definitions, to enable users to interpret each indicator independently without requiring external references. The JSON output reflects key dataset characteristics such as the number of samples, the feature types, the missingness levels, and the detected outliers, as well as global metrics (e.g., correlation differences, covariance similarity, mutual information difference, embedding-based scores) and per-feature local metrics such as KS, JS, WD, TVD, CV, and CC.

Furthermore, the module generates a rich suite of visual diagnostics that illustrate how real and synthetic datasets compare across statistical, structural, and topological axes. For distribution-level evaluation, the module generates histograms, kernel density estimation plots, and categorical frequency charts for each feature, enabling visual inspection of discrepancies in numeric distributions or category balance. Dependency-level plots include correlation heatmaps, covariance scatterplots, pairplot overlays, distance distribution comparisons, and association bar charts for categorical variables. These visual outputs facilitate the identification of mismatches in inter-feature dependencies and potential anomalies that may not be visible from metrics alone. For the structural and topological layers of the assessment, the module visualizes results from dimensionality-reduction embeddings via PCA, UMAP, and t-SNE, providing intuitive views of cluster alignments and density shifts between real and synthetic data. Additionally, it generates side-by-side k-nearest-neighbor graph visualizations based on PCA-reduced coordinates, providing qualitative insights into differences in local manifold geometry. These plots are automatically stored in a dedicated folder named after the "run_id", to facilitate comparisons across multiple synthetic data generation runs or model versions.

## 2.7. Extensibility and modularity

The SDB framework has been designed with extensibility and modularity to enable new evaluation components, metrics, or visualization elements to be integrated without altering existing functionalities.

Each module operates in an independent way with clearly defined interfaces. This modular design allows developers and researchers to extend or replace individual components while maintaining compatibility with the overall system pipeline. At the architectural level, extensibility is facilitated by separating metric computation, visualization, and reporting into distinct functional units. For example, new statistical metrics can be added by simply implementing an additional function inside the "metrics.py" module and returning the result as part of the existing per-feature or global metrics dictionary. The main workflow can automatically incorporate any newly defined metric, provided that it adheres to the standard input-output structure (e.g., functions that accept real and synthetic arrays and return scalar values). Similarly, dependency-level metrics can be augmented with alternative association measures without affecting other modules. The embedding-based and graph-based fidelity metrics, implemented in the "embeddings_graph.py", have been also structured for pluggability. If the embedding builder returns two matrices of equal dimensionality, all neighborhood, CKA, spectral, and Wasserstein-based metrics can operate flawlessly. New topology-aware metrics, message-passing graph kernel similarity, or node ranking consistency, can also be inserted into the same wrapper function. The reporting and visualization module also automatically adapts to new metrics when added to the global metrics dictionary. The SDB framework supports simple extensibility via optional arguments and modular plotting functions. The developers can easily integrate new plot types (e.g., latent density maps, feature drift timelines, or fairness disparity charts) into the "plot_utils.py", and they will be rendered into the run-specific output directory without changes to the main workflow. Because each plotting function takes dataset slices and configuration parameters independently, the plots are treated as composable building blocks. Finally, the JSON report dynamically incorporates new keys when additional metrics are added. The plots directory follows a consistent namespace defined by the unique "run_id", allowing parallel experiments or evaluation configurations to coexist without conflict.

## 3. Results

The SDB framework was evaluated across three diverse real-world domains, including healthcare, socioeconomic/financial modelling, and cybersecurity. These domains were selected to ensure broad coverage of variable types, fidelity challenges, and application-driven requirements. A use case (UC) was configured for each domain with a dedicated "config.yaml" file and executed through the unified SDB pipeline to perform feature-type detection, distribution-level fidelity assessment, dependency-preservation analysis, structural and embedding-based comparisons, outlier estimation, and completeness evaluation. The following subsections describe each use case in detail, incorporating empirical characteristics extracted from the JSON outputs (e.g., feature counts, outlier proportions, global correlations) to provide an accurate and methodologically sound interpretation. For demonstration purposes, the Bayesian Gaussian Mixture Models with Optimal Component Estimation (BGMMOCE) generator [20] was employed to synthesize datasets that match the original sample size in each UC, as it provides fast and reliable synthetic data generation across heterogeneous data types.

### 3.1. UC1 - Healthcare

UC1 focuses on a healthcare diagnostic setting using the open-source UCI Pima Indians Diabetes dataset [21]. The dataset includes 768 real and 768 synthetic samples, each containing nine features: seven continuous physiological variables (e.g., glucose, blood pressure, insulin, BMI), one ordinal feature (Pregnancies), and one binary categorical outcome, as automatically classified through the SDB's feature-type detection module. Data completeness is 100%, with no missing values, while outlier proportions among numerical variables range from minimal to moderate (e.g., 0.65% in glucose, 5.86% in blood pressure, 4.43% in insulin), all within clinically plausible ranges. Preserving clinically meaningful distributions and variable relationships is essential for UC1. The SDB's evaluation shows that the synthetic generator performs strongly in this regard: global dependency preservation is high, with low deviations in Pearson (0.028) and Spearman (0.033) correlations, indicating well-preserved inter-feature structure. Feature-level fidelity metrics in Table 2 confirm this behavior. Most continuous variables exhibit low KS statistics, such as glucose (0.06), blood pressure (0.03), and BMI (0.05), demonstrating close alignment between real and synthetic distributions. The ordinal variable Pregnancies also shows low divergence (KS = 0.05), suggesting that the generator effectively captured the discrete reproductive history pattern present in the population. As expected in medical datasets with sparsity or heavy tails, insulin (KS = 0.23) and skin thickness (KS = 0.17) show higher differences, which is consistent with their elevated Hellinger distances (0.15–0.19) and KLD values. Nevertheless, core physiological features such as BMI and blood pressure remain well preserved. Outcome-level fidelity is also strong. The binary label Outcome exhibits near-zero JSD (0.00), KLD (0.00), and very low TVD (0.02), while Cramér's V = 0.05 confirms minimal distortion in the association structure between the synthetic data and the target variable. This is critical for downstream predictive modelling, as it ensures the label distribution and its dependencies remain intact without compromising privacy. Overall, UC1 offers a robust benchmark for validating both distribution-level alignment and dependency-preserving fidelity of synthetic data generators in the healthcare domain.

**Table 2.** An instance of the metrics report for UC1.

| Variable [20] | Data type | KS | JSD | KLD | HD | TVD | CV | CC |
|---|---|---|---|---|---|---|---|---|
| Pregnancies | Ordinal | 0,05 | 0,01 | 0,09 | 0,12 | 0,14 | - | - |
| Glucose | Continuous | 0,06 | 0,02 | 0,16 | 0,14 | 0,12 | - | - |
| BloodPressure | Continuous | 0,03 | 0,03 | 0,20 | 0,18 | 0,16 | - | - |
| SkinThickness | Continuous | 0,17 | 0,03 | 0,15 | 0,19 | 0,16 | - | - |
| Insulin | Continuous | 0,23 | 0,02 | 0,23 | 0,15 | 0,12 | - | - |
| BMI | Continuous | 0,05 | 0,02 | 0,15 | 0,17 | 0,11 | - | - |
| DiabetesPedigreeFunction | Continuous | 0,12 | 0,05 | 0,44 | 0,25 | 0,24 | - | - |
| Age | Continuous | 0,12 | 0,07 | 0,34 | 0,30 | 0,27 | - | - |
| Outcome | Binary categorical | - | 0,00 | 0,00 | 0,02 | 0,02 | 0,05 | 1 |

A more granular inspection of the pairwise relationships among the variables is shown in Figure 2A, where scatterplot matrices and overlaid univariate KDEs illustrate a strong agreement between the real and synthetic samples across variables such as glucose, BMI, and blood pressure. Complementary univariate density comparisons for each continuous variable are presented in Figure 2B, again revealing close alignment, with slightly larger deviations in insulin and skin thickness, expected due to their skewed and sparse nature in the source population. Feature-level bivariate fidelity for "Diabetes Pedigree Function" and "Age" is highlighted in Figure 2C, where synthetic samples accurately replicate the distribution and joint relationships of the original dataset. Categorical feature fidelity is summarized in Figure 2D–E. The CV estimates demonstrate minimal dependency distortion, while the bar plots comparing the distributions of the "Outcome" and "Pregnancies" variables confirm that the synthetic generator preserved categorical proportion structures without mode collapse or label drift.

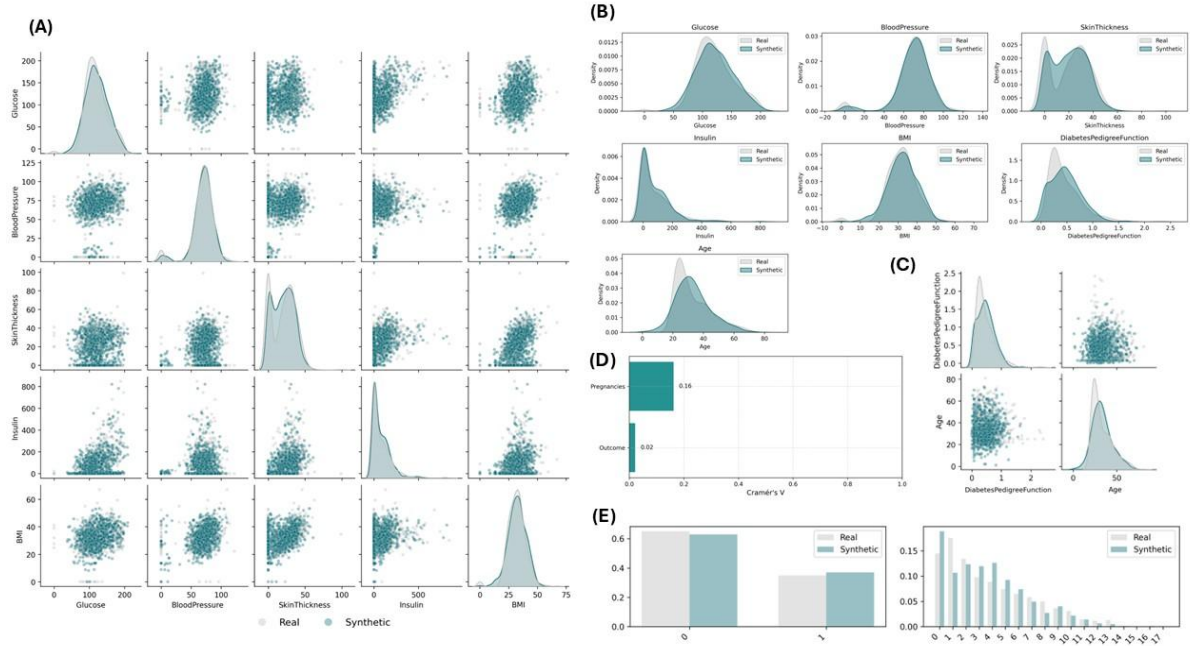

**Figure 2.** Distribution-level plots for UC1. (A) Scatterplot matrix with marginal KDEs comparing real (grey) and synthetic (teal) samples across continuous features. (B) Univariate kernel density estimates for individual numerical variables. (C) Continuation of (A) for the two remaining continuous features. (D) Cramér's V dependency scores for categorical variables. (E) Bar plots comparing categorical distributions (Outcome and Pregnancies) in real and synthetic data.

To further examine the global structure preservation between the real and synthetic data, we applied dimensionality reduction and graph-based analyses. Figure 3A illustrates the PCA projections which reveal consistent spread and orientation of real and synthetic data. Non-linear manifold embeddings using UMAP and t-SNE (Figure 3B, Figure 3C) confirm that the synthetic samples follow the same cluster formations and density regions as the real dataset.

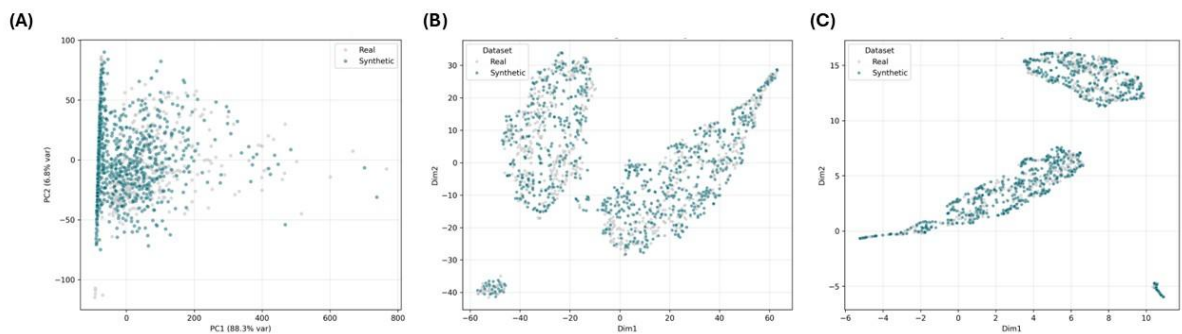

**Figure 3.** Embedding-based plots for UC1 (in 2D). (A) PCA projection of real and synthetic samples. (B) UMAP embedding showing cluster overlap. (C) t-SNE embedding showing preservation of manifold geometry.

To further evaluate the distributional fidelity and structural similarity between the real and synthetic data in UC1, we examined both correlation matrices and feature-level density relationships. Figure 4A, Figure 4B presents the Pearson correlation heatmaps for the real and synthetic data, which demonstrate close alignment across all numerical variables. The synthetic data preserve the weak-to-moderate inter-feature dependencies observed in the clinical dataset, which is consistent with the low Pearson and Spearman deviations reported earlier.

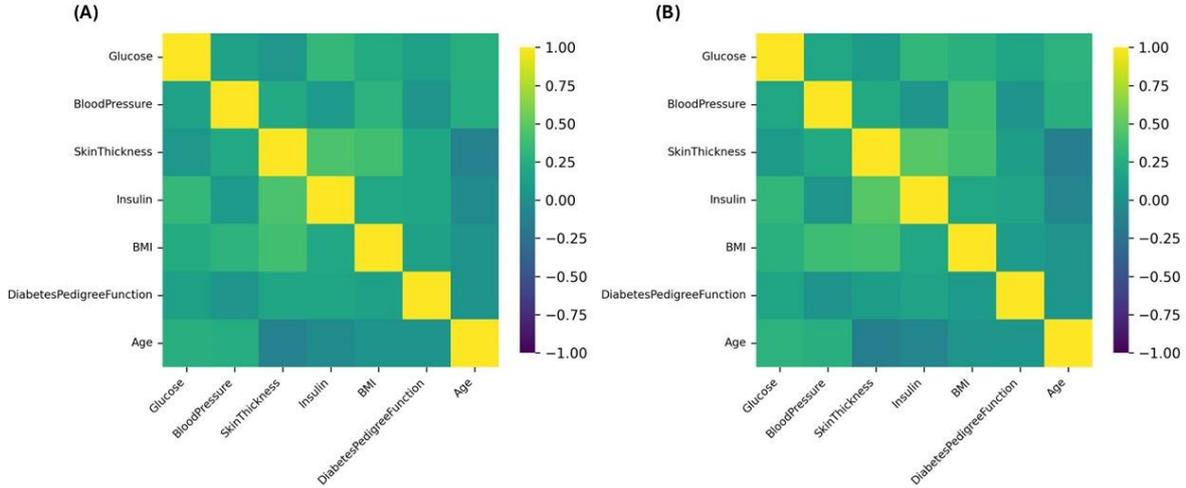

**Figure 4.** Dependency-based plots for UC1. (A) Pearson correlation matrix for the real dataset. (B) Pearson correlation matrix for the synthetic data.

## 3.2. UC2 - Socioeconomic / Finance

UC2 evaluates the SDB framework in a large-scale socioeconomic and financial modeling scenario using the open-source UCI Adult Income dataset [22]. This use case is considerably more complex than UC1, consisting of 39,215 real and 39,215 synthetic samples, with a mixture of continuous, ordinal, binary categorical, and multi-categorical variables. The feature set includes one ordinal variable ("education-num"), two binary categorical variables ("sex", "class"), several high-cardinality multi-categorical variables (e.g., "workclass", "occupation", "marital-status", "native-country"), and a smaller number of continuous variables. Many of these categorical features are highly imbalanced and contain long-tailed distributions, making UC2 an excellent testbed for evaluating the SDB's ability to preserve complex socio-demographic patterns. The dataset exhibits 100% completeness, with no missing values. Outlier rates vary from negligible to moderate depending on feature sparsity and tail behavior. This heterogeneity, combined with scale, poses significant challenges for synthetic data generation, particularly for preserving dependency structure across socioeconomic variables. Despite this complexity, the SDB achieved strong preservation of global structure. The Correlation Matrix Distance (CMD = 0.02), along with minimal Pearson (0.01) and Spearman (0.02) differences, indicates that linear and rank-based associations were well maintained across demographic and financial attributes. The Mutual Information Difference (MID = 0.03) further demonstrates consistent retention of nonlinear relationships, such as those linking "education" with "occupation", or "marital-status" with "class". Feature-level results from Table 3 provide deeper insight into dataset complexity. Many commonly represented demographic variables—such as "race", "sex", "relationship", "capitalgain", "capitalloss", and "hoursperweek"—exhibit very low divergence metrics, including near-zero JSD and KLD for "sex" and "capitalloss", and low TVD for "class" (0.02). These results demonstrate that major socio-demographic distributions were faithfully reproduced. In contrast, variables with sparse or long-tailed distributions show higher divergence, as expected. For example, "native-country" presents the highest divergence (JSD = 0.23; KLD = 0.78), reflecting its substantial category imbalance. Likewise, "education" and "education-num" show KS values between 0.19 and 0.20, indicating moderate distributional discrepancies. Nevertheless, all CC values equal 1, meaning that all real categories were preserved in the synthetic dataset, even where distributional alignment was imperfect. A notable example is "fnlwgt", a continuous variable with a heavy-tailed socioeconomic weighting distribution. It exhibits moderate KS (0.09) and JSD (0.03) values and a dependence score (CV = 0.50), illustrating the inherent difficulty of modeling highly skewed population-weighting variables. Conversely, "age" shows perfect categorical alignment (JSD = 0, KLD = 0, CC = 1) and low divergence values, confirming strong fidelity in stable demographic features. Given the Adult dataset's importance for fairness, socioeconomic analysis, and bias detection research, UC2 provides a stringent benchmark for evaluating whether synthetic data preserve representative socioeconomic patterns without amplifying group disparities. The results indicate that the SDB successfully maintained both global structural fidelity and local distributional coherence in a high-dimensional, category-rich dataset. The dataset's scale underscores the value of embedding-based and graph-based metrics for assessing structural consistency under complex categorical interactions.

**Table 3.** An instance of the metrics report for UC2.

| Variable [21] | Data type | KS | JSD | KLD | HD | TVD | CV | CC |
|---|---|---|---|---|---|---|---|---|
| workclass | Multi-categorical | - | 0,16 | 0,59 | 0,45 | 0,45 | 0,02 | 1 |
| fnlwgt | Continuous | 0,09 | 0,03 | 0,12 | 0,19 | 0,16 | - | - |
| education | Continuous | 0,19 | 0,16 | 0,65 | 0,42 | 0,49 | - | - |
| education-num | Ordinal | 0,20 | 0,10 | 0,38 | 0,32 | 0,39 | - | - |
| marital-status | Multi-categorical | - | 0,15 | 0,52 | 0,43 | 0,42 | 0,01 | 1 |
| occupation | Continuous | 0,14 | 0,11 | 0,40 | 0,36 | 0,37 | - | - |
| relationship | Multi-categorical | - | 0,07 | 0,23 | 0,26 | 0,26 | 0,01 | 1 |
| race | Multi-categorical | - | 0,01 | 0,05 | 0,12 | 0,06 | 0,01 | 1 |
| sex | Binary categorical | - | 0,00 | 0,00 | 0,02 | 0,03 | 0,01 | 1 |
| capitalgain | Multi-categorical | - | 0,07 | 0,25 | 0,27 | 0,25 | 0,01 | 1 |
| capitalloss | Multi-categorical | - | 0,00 | 0,00 | 0,02 | 0,01 | 0,01 | 1 |
| hoursperweek | Multi-categorical | - | 0,02 | 0,08 | 0,15 | 0,18 | 0,01 | 1 |
| native-country | Continuous | 0,46 | 0,23 | 0,78 | 0,53 | 0,54 | - | - |
| class | Binary categorical | - | 0,00 | 0,00 | 0,02 | 0,02 | 0,00 | 1 |

### 3.3. UC3 - Cybersecurity / Network Traffic

UC3 evaluates the SDB framework within the cybersecurity and network traffic domain using the InSDN Intrusion Detection dataset [23], a high-dimensional real-world telemetry dataset widely used in intrusion detection research. It contains 109,394 real samples and an equal number of synthetic samples, each described by 65+ flow-based statistical features. The dataset includes predominantly continuous variables—such as inter-arrival times, packet-length metrics, header-byte statistics, subflow indicators, and flow-level rate measurements, alongside several binary and multi-categorical attributes (e.g., protocol identifiers, TCP flag counts, and attack labels). Feature-type detection identified 54 continuous variables, five binary categorical variables, and three multi-categorical variables, confirming that UC3 is highly heterogeneous yet overwhelmingly numeric. This domain is characterized by heavy-tailed, bursty, and non-stationary distributions, reflecting realistic network behavior. Such statistical properties naturally lead to high divergence scores in rare-event or extreme-value features when compared to synthetic approximations. The distributional fidelity results in Table 4 show domain-consistent patterns: features representing extremes, spikes, or highly skewed values exhibit elevated divergences. For example, "Flow IAT Min" shows KS = 0.49, JSD = 0.58, and KLD = 4.20, while "Init Bwd Win Byts" exhibits JSD = 0.60 and KLD = 4.13. Similar behavior is observed in "Bwd IAT Min" (JSD = 0.44, KLD = 3.12). These large divergences are expected because tail-heavy variables encode unique semantic fingerprints of network flows and are particularly sensitive to generator variability. Packet-length related features—including "Fwd Pkt Len Mean", "Bwd Pkt Len Std", "Pkt Len Mean", and "Fwd Seg Size Avg"—show KS ≈ 0.42–0.43 and KLD values exceeding 1.0, reflecting known challenges in learning packet-size variability arising from protocol constraints, application behaviors, and microburst events. Despite these deviations, the generator still maintains coherent distributional structure across stable flow metrics (e.g., "Tot Fwd Pkts", "Flow Duration"). In contrast, categorical fidelity is exceptionally strong.

Protocol identifiers such as "Protocol" achieve near-perfect alignment (JSD = 0.00, KLD = 0.00, CC = 1.00). Similarly, TCP flag indicators—"FIN Flag Cnt", "SYN Flag Cnt", "PSH Flag Cnt", and "ACK Flag Cnt"—exhibit JSD < 0.01, KLD < 0.04, and CC = 1.00. These binary forensic features are critical because they define the semantic structure of many intrusion-detection signatures. Preserving them with such high fidelity ensures that synthetic data remain meaningful for downstream IDS evaluation. The attack label variable "Label" also shows good categorical consistency (JSD = 0.20, KLD = 0.65, CC = 1.00). Although the JSD and KLD values are moderate—reflecting distributional differences across benign vs. attack classes—the perfect CC score confirms that all real-world label categories are represented in the synthetic data. This is essential for ensuring that synthetic intrusion datasets remain usable for model training and benchmark evaluation. Given UC3's high dimensionality and the intricate dependencies among flow-based features, this use case is particularly valuable for stress-testing SDB's structural fidelity metrics. In such environments, capturing manifold geometry, cluster relationships, and high-order interactions is just as important as univariate distributional matching. Metrics such as k-nearest neighbor (kNN) neighborhood overlap, SD, and AWED scores are crucial for verifying whether synthetic data preserve relational structure across behavioral clusters, e.g., benign browsing traffic, DoS floods, TCP scans, or protocol-specific subflows. Finally, cybersecurity datasets inherently contain unique behavioral fingerprints that could threaten privacy if memorized by a generator. The high divergences in rare or bursty variables, such as inter-arrival minima, window sizes, and packet-length maxima, are thus reassuring signs that the synthetic data avoid memorizing exact traffic patterns, supporting both privacy protection and distributional safety. UC3 therefore serves as a rigorous benchmark for evaluating synthetic data quality in complex, protocol-driven, privacy-sensitive environments.

**Table 4.** An instance of the metrics report for UC3.

| Variable [22] | Data type | KS | JSD | KLD | HD | TVD | CV | CC |
|---|---|---|---|---|---|---|---|---|
| Src Port | Continuous | 0,47 | 0,19 | 0,66 | 0,47 | 0,49 | - | - |
| Dst Port | Continuous | 0,35 | 0,16 | 0,54 | 0,45 | 0,43 | - | - |
| Protocol | Multi-categorical | - | 0,00 | 0,00 | 0,02 | 0,02 | 0 | 1 |
| Flow Duration | Continuous | 0,41 | 0,22 | 0,85 | 0,51 | 0,52 | - | - |
| Tot Fwd Pkts | Continuous | 0,20 | 0,13 | 0,44 | 0,40 | 0,34 | - | - |
| Tot Bwd Pkts | Continuous | 0,31 | 0,23 | 0,80 | 0,53 | 0,53 | - | - |
| TotLen Fwd Pkts | Continuous | 0,36 | 0,26 | 0,91 | 0,59 | 0,57 | - | - |
| TotLen Bwd Pkts | Continuous | 0,40 | 0,27 | 0,94 | 0,60 | 0,60 | - | - |
| Fwd Pkt Len Max | Continuous | 0,40 | 0,27 | 0,93 | 0,60 | 0,60 | - | - |
| Fwd Pkt Len Mean | Continuous | 0,43 | 0,34 | 1,46 | 0,67 | 0,66 | - | - |
| Fwd Pkt Len Std | Continuous | 0,43 | 0,32 | 1,35 | 0,65 | 0,65 | - | - |
| Bwd Pkt Len Max | Continuous | 0,40 | 0,27 | 0,94 | 0,60 | 0,60 | - | - |
| Bwd Pkt Len Mean | Continuous | 0,42 | 0,33 | 1,41 | 0,65 | 0,65 | - | - |
| Bwd Pkt Len Std | Continuous | 0,43 | 0,32 | 1,38 | 0,64 | 0,65 | - | - |
| Flow Byts/s | Continuous | 0,40 | 0,16 | 0,53 | 0,46 | 0,41 | - | - |
| Flow Pkts/s | Continuous | 0,25 | 0,12 | 0,49 | 0,40 | 0,29 | - | - |
| Flow IAT Mean | Continuous | 0,42 | 0,23 | 0,90 | 0,52 | 0,53 | - | - |
| Flow IAT Std | Continuous | 0,43 | 0,26 | 1,25 | 0,56 | 0,56 | - | - |
| Flow IAT Max | Continuous | 0,43 | 0,25 | 1,15 | 0,54 | 0,55 | - | - |
| Flow IAT Min | Continuous | 0,49 | 0,58 | 4,20 | 0,88 | 0,90 | - | - |
| Fwd IAT Tot | Continuous | 0,40 | 0,17 | 0,67 | 0,47 | 0,41 | - | - |
| Fwd IAT Mean | Continuous | 0,41 | 0,16 | 0,65 | 0,45 | 0,42 | - | - |
| Fwd IAT Std | Continuous | 0,42 | 0,17 | 0,67 | 0,47 | 0,43 | - | - |
| Fwd IAT Max | Continuous | 0,41 | 0,17 | 0,69 | 0,47 | 0,42 | - | - |
| Fwd IAT Min | Continuous | 0,38 | 0,12 | 0,51 | 0,37 | 0,39 | - | - |
| Bwd IAT Tot | Continuous | 0,41 | 0,23 | 1,03 | 0,53 | 0,53 | - | - |
| Bwd IAT Mean | Continuous | 0,42 | 0,25 | 1,26 | 0,55 | 0,54 | - | - |
| Bwd IAT Std | Continuous | 0,42 | 0,25 | 1,25 | 0,56 | 0,55 | - | - |
| Bwd IAT Max | Continuous | 0,42 | 0,24 | 1,21 | 0,54 | 0,54 | - | - |
| Bwd IAT Min | Continuous | 0,40 | 0,44 | 3,12 | 0,74 | 0,79 | - | - |
| Bwd PSH Flags | Multi-categorical | - | 0,01 | 0,04 | 0,11 | 0,04 | 0 | 1 |
| Fwd Header Len | Continuous | 0,41 | 0,25 | 1,03 | 0,58 | 0,55 | - | - |
| Bwd Header Len | Continuous | 0,31 | 0,27 | 1,10 | 0,59 | 0,58 | - | - |
| Fwd Pkts/s | Continuous | 0,41 | 0,19 | 0,63 | 0,50 | 0,45 | - | - |
| Bwd Pkts/s | Continuous | 0,22 | 0,13 | 0,52 | 0,41 | 0,31 | - | - |
| Pkt Len Max | Continuous | 0,40 | 0,28 | 0,94 | 0,60 | 0,60 | - | - |
| Pkt Len Mean | Continuous | 0,42 | 0,32 | 1,31 | 0,65 | 0,65 | - | - |
| Pkt Len Std | Continuous | 0,43 | 0,31 | 1,14 | 0,63 | 0,64 | - | - |
| Pkt Len Var | Continuous | 0,42 | 0,24 | 1,01 | 0,56 | 0,51 | - | - |
| FIN Flag Cnt | Multi-categorical | - | 0,00 | 0,00 | 0,03 | 0,04 | 0 | 1 |
| SYN Flag Cnt | Multi-categorical | - | 0,00 | 0,01 | 0,06 | 0,07 | 0 | 1 |
| PSH Flag Cnt | Multi-categorical | - | 0,01 | 0,04 | 0,11 | 0,04 | 0 | 1 |
| ACK Flag Cnt | Multi-categorical | - | 0,00 | 0,00 | 0,03 | 0,04 | 0 | 1 |
| Down/Up Ratio | Multi-categorical | - | 0,02 | 0,08 | 0,17 | 0,09 | 0 | 1 |
| Pkt Size Avg | Continuous | 0,42 | 0,32 | 1,29 | 0,64 | 0,64 | - | - |
| Fwd Seg Size Avg | Continuous | 0,43 | 0,34 | 1,46 | 0,67 | 0,66 | - | - |
| Bwd Seg Size Avg | Continuous | 0,42 | 0,33 | 1,41 | 0,65 | 0,65 | - | - |
| Subflow Fwd Pkts | Continuous | 0,20 | 0,14 | 0,44 | 0,41 | 0,35 | - | - |
| Subflow Fwd Byts | Continuous | 0,36 | 0,26 | 0,90 | 0,59 | 0,57 | - | - |
| Subflow Bwd Pkts | Continuous | 0,31 | 0,23 | 0,80 | 0,53 | 0,53 | - | - |
| Subflow Bwd Byts | Continuous | 0,40 | 0,27 | 0,94 | 0,60 | 0,60 | - | - |
| Init Bwd Win Byts | Continuous | 0,37 | 0,60 | 4,13 | 0,90 | 0,94 | - | - |
| Fwd Act Data Pkts | Continuous | 0,08 | 0,00 | 0,00 | 0,02 | 0,00 | - | - |
| Active Mean | Continuous | 0,47 | 0,20 | 0,68 | 0,51 | 0,49 | - | - |
| Active Std | Continuous | 0,50 | 0,21 | 0,71 | 0,54 | 0,50 | - | - |
| Active Max | Continuous | 0,47 | 0,20 | 0,69 | 0,51 | 0,49 | - | - |
| Active Min | Continuous | 0,47 | 0,20 | 0,68 | 0,51 | 0,49 | - | - |
| Idle Mean | Continuous | 0,46 | 0,27 | 1,28 | 0,59 | 0,58 | - | - |
| Idle Std | Continuous | 0,50 | 0,22 | 0,72 | 0,54 | 0,51 | - | - |
| Idle Max | Continuous | 0,46 | 0,27 | 1,26 | 0,59 | 0,57 | - | - |
| Idle Min | Continuous | 0,46 | 0,27 | 1,29 | 0,59 | 0,58 | - | - |
| Label | Multi-categorical | - | 0,20 | 0,65 | 0,48 | 0,50 | 0 | 1 |

## 4. Discussion

The results across the three use cases confirm that the SDB provides a robust, comprehensive, and interpretable framework for evaluating synthetic tabular data across heterogeneous domains. Unlike existing evaluation practices that focus on narrow subsets of metrics or rely on ad-hoc visual comparisons, SDB demonstrates the value of a unified, multi-layered methodology that jointly examines statistical fidelity, dependency preservation, structural alignment in embedding space, and graph-theoretic consistency. The findings highlight that no single metric is sufficient for validating synthetic data. Instead, fidelity must be assessed along multiple axes; distributional, relational, and topological, to obtain a complete understanding of generator behavior and potential risks. A key outcome that emerges from the three use cases lies on the importance of domain-aware interpretation of fidelity metrics.

In UC1, the SDB revealed that synthetic data replicated physiologically meaningful distributions with high precision, particularly for well-behaved variables such as glucose, BMI, and blood pressure. Higher divergences in insulin and skin thickness were appropriately contextualized as being driven by sparsity and natural skewness in the original population. These patterns illustrate that synthetic data evaluation benefits from grounding in domain knowledge, particularly when interpreting deviations that are intrinsic to the real dataset rather than indicative of generator failure. UC2 highlighted SDB's ability to handle high-cardinality categorical features, which are often overlooked in traditional evaluation pipelines. Despite substantial heterogeneity and rare categories, especially in occupation, native-country, and workclass, the synthetic data preserved category membership and mutual information patterns, demonstrating that fidelity assessment for categorical data must incorporate both distributional alignment and dependency-aware metrics. The preservation of Pearson, Spearman, and mutual-information structures emphasizes that synthetic data can retain meaningful socio-demographic relationships without replicating sensitive individual-level details, if evaluation is sufficiently holistic.

UC3 illustrates SDB's ability to operate effectively on high-dimensional, heavy-tailed, bursty, and protocol-governed datasets. In such contexts, elevated divergences in packet-length and inter-arrival distributions are expected and reflect intrinsic irregularities in real traffic rather than structural faults in the synthetic generator. Importantly, categorical fidelity across protocol identifiers and TCP flag features was nearly perfect, illustrating that SDB can reveal which aspects of a dataset are faithfully reproduced (e.g., event-type structure) versus which reflect unavoidable uncertainty (e.g., rare extreme values). Moreover, the embedding- and graph-based components proved essential for the diagnosis of manifold-level preservation, a critical requirement for intrusion detection research, where synthetic flows must reflect realistic cluster structures without risking reconstruction of unique behavioral signatures. Across the three domains, a common observation is that synthetic data fidelity is intrinsically multidimensional, and deviations must be interpreted through the combined lens of statistical, structural, dependency-based, and privacy-aware metrics. SDB's modular pipeline provides such an integrated perspective, revealing not only how real and synthetic datasets differ but also why these differences arise and whether they are acceptable for downstream tasks. The results demonstrate that relying solely on univariate tests such as KS or $\chi^2$ can lead to incomplete or misleading conclusions, particularly in datasets with complex dependency structures or non-Gaussian behavior. The embedding- and graph-based metrics in SDB capture aspects of data geometry that would otherwise remain invisible to purely statistical evaluations.

An additional contribution of SDB is its emphasis on transparency and reproducibility. The structured JSON reports, systematic naming conventions, and automated visual outputs provide a standardized foundation for auditing synthetic data quality. This is particularly important for critical domains, like healthcare, finance, and cybersecurity, where synthetic data may be used for model development, what-if simulations, or fairness auditing. By producing a complete, machine-readable record of fidelity, SDB supports traceability and aligns with emerging regulatory requirements under frameworks such as GDPR, EU AI Act, and sector-specific governance standards. SDB also highlights ongoing challenges in synthetic data science. Feature-level divergences may be driven by limitations in the generator, intrinsic dataset irregularities, or imbalance patterns that are difficult to model. Embedding-based evaluation introduces dependence on representation models, which may vary in quality, especially for highly categorical datasets. Furthermore, while SDB includes distance-based privacy risk indicators, future extensions could integrate formal privacy metrics such as differential privacy bounds, membership inference attack scores, or reconstruction risk indexes. These represent promising directions to improve privacy-aware fidelity assessment.

# 5. Conclusions and Future Work

Our findings demonstrate that SDB is a domain-agnostic framework capable of evaluating synthetic data across diverse real-world conditions. SDB provides a straightforward methodological foundation to ensure the trustworthiness, transparency, and practical usability of synthetic tabular datasets by unifying statistical, structural, and privacy-related components within a modular and extensible pipeline. The three case studies highlight that SDB can reliably assess fidelity across heterogeneous feature types, varying scales, and domain-specific statistical challenges. This positions SDB as a valuable framework not only for research and model benchmarking, but also for industrial deployment and regulatory auditing, where synthetic data must be validated with methodological consistency and interpretability. Looking ahead, several directions will further enhance SDB's applicability and automation capabilities. First, we are actively developing enriched automated PDF reporting functionality, which will integrate structured narratives, per-metric interpretations, feature-level summaries, and embedded visual diagnostics. This will allow practitioners to generate publication-grade evaluation reports with a single command, thus strengthening SDB's role in auditability and compliance contexts. Second, we plan to extend SDB's core architecture to support additional data modalities, including time-series data, longitudinal records, graph-structured data, and multimodal tabular–text hybrids. These extensions will broaden SDB's relevance to emerging application domains such as wearable sensor analytics, clinical longitudinal modelling, financial forecasting, and cyber-physical systems. Third, future work will focus on the validation of the SDB across a wider spectrum of domains and generative models, such as federated synthetic data generation [23], differentially private mechanisms [24], diffusion-based tabular models [25], and foundation-model-driven synthetic data frameworks [26], to ensure that the blueprint remains aligned with the rapidly evolving landscape of generative AI. Finally, we aim to integrate more explicit privacy risk quantification tools, including adversarial membership inference tests, nearest-neighbor memorization indicators, and domain-aware privacy scorecards, complementing SDB's existing distance-based privacy diagnostics.

# Appendix

**The complete metrics report from UC1 (.JSON file)**

```
{
  "metadata": {
    "run_id": "sdb_978019ef05af",
    "timestamp": "2025-12-10T17:32:37.389757",
```

```
            "real_dataset_path": "test/UC1_real.csv",
            "synthetic_dataset_path": "test/UC1_synth.csv",
            "number_of_samples_real": 768,
            "number_of_samples_synthetic": 768,
            "total_features": 9,
            "numerical_features": 8,
            "binary_categorical_features": 1,
            "multi_categorical_features": 0,
            "total_missing_values": 0,
            "data_completeness (%)": 100.0,
            "outliers (%)": {
                "Pregnancies": 0.52,
                "Glucose": 0.65,
                "BloodPressure": 5.86,
                "SkinThickness": 0.13,
                "Insulin": 4.43,
                "BMI": 2.47,
                "DiabetesPedigreeFunction": 3.78,
                "Age": 1.17
            }
        },
    "metric_definitions": {
        "Kolmogorov–Smirnov (KS) Statistic": "Measures the maximum distance between the empirical cumulative distributions of real and synthetic data for a numeric feature.",
        "Kullback-Leibler Divergence (KLD)": "Quantifies how much information is lost when approximating the real data distribution with the synthetic one. Asymmetric measure.",
        "Jensen–Shannon (JS) Divergence (JSD)": "Symmetric measure of similarity between two probability distributions derived from real and synthetic data. Lower values indicate higher similarity.",
        "Wasserstein Distance (WD)": "Quantifies the minimum 'work' required to transform one probability distribution into another, reflecting both shape and distance differences.",
        "Hellinger Distance (HD)": "Measures the distance between two probability distributions; bounded between 0 (identical) and 1 (completely dissimilar).",
        "Total Variation Distance (TVD)": "Measures the maximum absolute difference between two probability distributions. Values range from 0 (identical) to 1 (completely disjoint). Supports both numeric and categorical data.",
        "Range Coverage (RC)": "Fraction of the real data's numeric range that is covered by the synthetic data. Values close to 1 indicate the synthetic data spans the same domain as the real data.",
        "Chi-Square Statistic (CSS)": "Tests whether the observed category frequencies in the synthetic data differ significantly from those in the real data.",
        "Category Coverage (CC)": "Proportion of unique categories in the real data that also appear in the synthetic data; detects missing or underrepresented categories.",
        "Contingency Table Similarity (CV)": "Measures the strength of association between two categorical variables in real vs. synthetic datasets; used to compare inter-feature dependencies.",
        "Covariance Matrix Similarity (CMS)": "Quantifies deviation between real and synthetic covariance matrices; smaller Frobenius norm indicates closer similarity.",
        "Correlation Matrix Distance (CMD)": "Computes normalized Frobenius norm of the difference between correlation matrices; used as an overall measure of structural fidelity.",
```

```
      "Correlation Difference (Pearson) (CDP)": "Measures how much the linear (Pearson) correlations between features differ between real and synthetic datasets.",
      "Correlation Difference (Spearman) (CDS)": "Measures how much the rank (Spearman) correlations between features differ between real and synthetic datasets.",
      "Mutual Information Difference (MID)": "Captures how well nonlinear dependencies between variables are preserved; compares mutual information matrices between real and synthetic data.",
      "Centered Kernel Alignment (CKA)": "Measures similarity between real and synthetic feature representations in embedding space. Values range from 0 (no similarity) to 1 (identical representation).",
      "Average Wasserstein Embedding Distance (AWED)": "Average Wasserstein distance between real and synthetic points in embedding space. Lower values indicate better alignment of sample distributions.",
      "Neighbor Overlap (Jaccard Similarity)": "Measures how similar each sample's nearest-neighbor set is between real and synthetic data. Calculated using Jaccard index between the kNN lists of real and synthetic embeddings.",
      "Spectral Distance (SD)": "Distance between the eigenvalue spectra of real and synthetic kNN graphs. Lower values indicate better preservation of global graph structure.",
      "Graph Structural Fidelity Score (GSFS)": "Measures global structural preservation of the kNN graph by comparing degree distributions, clustering coefficients, and shortest-path distances. Values range from 0 to 1, with higher values indicating better global topology preservation."
    },
    "global_metrics": {
      "Covariance_Matrix_Similarity_Frobenius": 2646.20781,
      "Correlation_Matrix_Distance": 0.08786,
      "Correlation_Difference_Pearson": 0.02838,
      "Correlation_Difference_Spearman": 0.03312,
      "Mutual_Information_Difference": null,
      "CKA": 0.01199,
      "Neighborhood_Overlap": 0.11289,
      "Spectral_Distance": 100.08007,
      "Avg_Wasserstein_Embedding": 0.02884,
      "GSFS": 0.73829
    },
    "local_metrics": {
      "Pregnancies": {
        "KS_Statistic": 0.05208,
        "JS_Divergence": 0.01367,
        "KL_Divergence": 0.09323,
        "Wasserstein_Distance": 0.3724,
        "Hellinger_Distance": 0.11972,
        "Total_Variation_Distance": 0.13672,
        "Range_Coverage": 0.94118,
        "Chi_Square_Statistic": null,
        "Contingency_CramerV": null,
        "Category_Coverage": null
      },
      "Glucose": {
        "KS_Statistic": 0.0599,
```

```
      "JS_Divergence": 0.01819,
      "KL_Divergence": 0.15765,
      "Wasserstein_Distance": 3.05208,
      "Hellinger_Distance": 0.14266,
      "Total_Variation_Distance": 0.1237,
      "Range_Coverage": 0.80905,
      "Chi_Square_Statistic": null,
      "Contingency_CramerV": null,
      "Category_Coverage": null
    },
    "BloodPressure": {
      "KS_Statistic": 0.02865,
      "JS_Divergence": 0.02752,
      "KL_Divergence": 0.20405,
      "Wasserstein_Distance": 1.17057,
      "Hellinger_Distance": 0.17981,
      "Total_Variation_Distance": 0.15885,
      "Range_Coverage": 0.90164,
      "Chi_Square_Statistic": null,
      "Contingency_CramerV": null,
      "Category_Coverage": null
    },
    "SkinThickness": {
      "KS_Statistic": 0.17057,
      "JS_Divergence": 0.03355,
      "KL_Divergence": 0.15005,
      "Wasserstein_Distance": 2.11589,
      "Hellinger_Distance": 0.19019,
      "Total_Variation_Distance": 0.16406,
      "Range_Coverage": 0.61616,
      "Chi_Square_Statistic": null,
      "Contingency_CramerV": null,
      "Category_Coverage": null
    },
    "Insulin": {
      "KS_Statistic": 0.23177,
      "JS_Divergence": 0.01864,
      "KL_Divergence": 0.22887,
      "Wasserstein_Distance": 10.79818,
      "Hellinger_Distance": 0.14796,
      "Total_Variation_Distance": 0.11589,
      "Range_Coverage": 0.96809,
      "Chi_Square_Statistic": null,
      "Contingency_CramerV": null,
      "Category_Coverage": null
    },
    "BMI": {
      "KS_Statistic": 0.05469,
```

```json
      "JS_Divergence": 0.02221,
      "KL_Divergence": 0.14632,
      "Wasserstein_Distance": 0.88746,
      "Hellinger_Distance": 0.16897,
      "Total_Variation_Distance": 0.10547,
      "Range_Coverage": 0.79485,
      "Chi_Square_Statistic": null,
      "Contingency_CramerV": null,
      "Category_Coverage": null
    },
    "DiabetesPedigreeFunction": {
      "KS_Statistic": 0.11979,
      "JS_Divergence": 0.05452,
      "KL_Divergence": 0.44246,
      "Wasserstein_Distance": 0.06496,
      "Hellinger_Distance": 0.25298,
      "Total_Variation_Distance": 0.23958,
      "Range_Coverage": 0.67162,
      "Chi_Square_Statistic": null,
      "Contingency_CramerV": null,
      "Category_Coverage": null
    },
    "Age": {
      "KS_Statistic": 0.1224,
      "JS_Divergence": 0.07399,
      "KL_Divergence": 0.34161,
      "Wasserstein_Distance": 1.94922,
      "Hellinger_Distance": 0.30314,
      "Total_Variation_Distance": 0.27214,
      "Range_Coverage": 0.86667,
      "Chi_Square_Statistic": null,
      "Contingency_CramerV": null,
      "Category_Coverage": null
    },
    "Outcome": {
      "KS_Statistic": null,
      "JS_Divergence": 0.00027,
      "KL_Divergence": 0.00106,
      "Hellinger_Distance": 0.01631,
      "Total_Variation_Distance": 0.02214,
      "Wasserstein_Distance": null,
      "Chi_Square_Statistic": 1.65636,
      "Contingency_CramerV": 0.05345,
      "Category_Coverage": 1.0,
      "Range_Coverage": null
    }
  }
}
```

**The complete metrics report from UC2 (.JSON file)**

```
{
  "metadata": {
    "run_id": "sdb_6f9c33225374",
    "timestamp": "2025-12-10T17:33:20.263405",
    "real_dataset_path": "test/UC2_real.csv",
    "synthetic_dataset_path": "test/UC2_synth.csv",
    "number_of_samples_real": 39215,
    "number_of_samples_synthetic": 39215,
    "total_features": 15,
    "numerical_features": 5,
    "binary_categorical_features": 2,
    "multi_categorical_features": 8,
    "total_missing_values": 0,
    "data_completeness (%)": 100.0,
    "outliers (%)": {
      "fnlwgt": 3.03,
      "education": 7.65,
      "education-num": 0.74,
      "occupation": 0.0,
      "native-country": 10.52
    }
  },
  "metric_definitions": {
    "Kolmogorov–Smirnov (KS) Statistic": "Measures the maximum distance between the empirical cumulative distributions of real and synthetic data for a numeric feature.",
    "Kullback-Leibler Divergence (KLD)": "Quantifies how much information is lost when approximating the real data distribution with the synthetic one. Asymmetric measure.",
    "Jensen–Shannon (JS) Divergence (JSD)": "Symmetric measure of similarity between two probability distributions derived from real and synthetic data. Lower values indicate higher similarity.",
    "Wasserstein Distance (WD)": "Quantifies the minimum 'work' required to transform one probability distribution into another, reflecting both shape and distance differences.",
    "Hellinger Distance (HD)": "Measures the distance between two probability distributions; bounded between 0 (identical) and 1 (completely dissimilar).",
    "Total Variation Distance (TVD)": "Measures the maximum absolute difference between two probability distributions. Values range from 0 (identical) to 1 (completely disjoint). Supports both numeric and categorical data.",
    "Range Coverage (RC)": "Fraction of the real data's numeric range that is covered by the synthetic data. Values close to 1 indicate the synthetic data spans the same domain as the real data.",
    "Chi-Square Statistic (CSS)": "Tests whether the observed category frequencies in the synthetic data differ significantly from those in the real data.",
    "Category Coverage (CC)": "Proportion of unique categories in the real data that also appear in the synthetic data; detects missing or underrepresented categories.",
    "Contingency Table Similarity (CV)": "Measures the strength of association between two categorical variables in real vs. synthetic datasets; used to compare inter-feature dependencies.",
```

```
    "Covariance Matrix Similarity (CMS)": "Quantifies deviation between real and synthetic
covariance matrices; smaller Frobenius norm indicates closer similarity.",
    "Correlation Matrix Distance (CMD)": "Computes normalized Frobenius norm of the difference
between correlation matrices; used as an overall measure of structural fidelity.",
    "Correlation Difference (Pearson) (CDP)": "Measures how much the linear (Pearson)
correlations between features differ between real and synthetic datasets.",
    "Correlation Difference (Spearman) (CDS)": "Measures how much the rank (Spearman)
correlations between features differ between real and synthetic datasets.",
    "Mutual Information Difference (MID)": "Captures how well nonlinear dependencies between
variables are preserved; compares mutual information matrices between real and synthetic data.",
    "Centered Kernel Alignment (CKA)": "Measures similarity between real and synthetic feature
representations in embedding space. Values range from 0 (no similarity) to 1 (identical
representation).",
    "Average Wasserstein Embedding Distance (AWED)": "Average Wasserstein distance between
real and synthetic points in embedding space. Lower values indicate better alignment of sample
distributions.",
    "Neighbor Overlap (Jaccard Similarity)": "Measures how similar each sample's nearest-neighbor
set is between real and synthetic data. Calculated using Jaccard index between the kNN lists of real
and synthetic embeddings.",
    "Spectral Distance (SD)": "Distance between the eigenvalue spectra of real and synthetic kNN
graphs. Lower values indicate better preservation of global graph structure.",
    "Graph Structural Fidelity Score (GSFS)": "Measures global structural preservation of the kNN
graph by comparing degree distributions, clustering coefficients, and shortest-path distances. Values
range from 0 to 1, with higher values indicating better global topology preservation."
  },
  "global_metrics": {
    "Covariance_Matrix_Similarity_Frobenius": 702615667.10881,
    "Correlation_Matrix_Distance": 0.01739,
    "Correlation_Difference_Pearson": 0.00645,
    "Correlation_Difference_Spearman": 0.02278,
    "Mutual_Information_Difference": 0.03276,
    "CKA": 0.00455,
    "Neighborhood_Overlap": 0.10381,
    "Spectral_Distance": 102.61173,
    "Avg_Wasserstein_Embedding": 0.03862,
    "GSFS": 0.69229
  },
  "local_metrics": {
    "age": {
      "KS_Statistic": null,
      "JS_Divergence": 0.0,
      "KL_Divergence": -0.0,
      "Hellinger_Distance": 0.0,
      "Total_Variation_Distance": 0.0,
      "Wasserstein_Distance": null,
      "Chi_Square_Statistic": null,
      "Contingency_CramerV": NaN,
      "Category_Coverage": 1.0,
```

```
      "Range_Coverage": null
    },
    "workclass": {
      "KS_Statistic": null,
      "JS_Divergence": 0.16338,
      "KL_Divergence": 0.58742,
      "Hellinger_Distance": 0.4452,
      "Total_Variation_Distance": 0.44876,
      "Wasserstein_Distance": null,
      "Chi_Square_Statistic": 76412535.65902,
      "Contingency_CramerV": 0.01567,
      "Category_Coverage": 1.0,
      "Range_Coverage": null
    },
    "fnlwgt": {
      "KS_Statistic": 0.0868,
      "JS_Divergence": 0.03157,
      "KL_Divergence": 0.12326,
      "Wasserstein_Distance": 13341.44483,
      "Hellinger_Distance": 0.19395,
      "Total_Variation_Distance": 0.16076,
      "Range_Coverage": 0.5032,
      "Chi_Square_Statistic": null,
      "Contingency_CramerV": null,
      "Category_Coverage": null
    },
    "education": {
      "KS_Statistic": 0.18669,
      "JS_Divergence": 0.16073,
      "KL_Divergence": 0.6498,
      "Wasserstein_Distance": 1.07043,
      "Hellinger_Distance": 0.42259,
      "Total_Variation_Distance": 0.48902,
      "Range_Coverage": 1.0,
      "Chi_Square_Statistic": null,
      "Contingency_CramerV": null,
      "Category_Coverage": null
    },
    "education-num": {
      "KS_Statistic": 0.20316,
      "JS_Divergence": 0.09972,
      "KL_Divergence": 0.37627,
      "Wasserstein_Distance": 0.63578,
      "Hellinger_Distance": 0.32423,
      "Total_Variation_Distance": 0.38531,
      "Range_Coverage": 1.0,
      "Chi_Square_Statistic": null,
      "Contingency_CramerV": null,
```

```
      "Category_Coverage": null
    },
    "marital-status": {
      "KS_Statistic": null,
      "JS_Divergence": 0.15455,
      "KL_Divergence": 0.51633,
      "Hellinger_Distance": 0.4251,
      "Total_Variation_Distance": 0.42384,
      "Wasserstein_Distance": null,
      "Chi_Square_Statistic": 1681028.71861,
      "Contingency_CramerV": 0.01366,
      "Category_Coverage": 1.0,
      "Range_Coverage": null
    },
    "occupation": {
      "KS_Statistic": 0.14127,
      "JS_Divergence": 0.11224,
      "KL_Divergence": 0.40181,
      "Wasserstein_Distance": 0.81716,
      "Hellinger_Distance": 0.36128,
      "Total_Variation_Distance": 0.36789,
      "Range_Coverage": 1.0,
      "Chi_Square_Statistic": null,
      "Contingency_CramerV": null,
      "Category_Coverage": null
    },
    "relationship": {
      "KS_Statistic": null,
      "JS_Divergence": 0.06594,
      "KL_Divergence": 0.23114,
      "Hellinger_Distance": 0.2644,
      "Total_Variation_Distance": 0.26069,
      "Wasserstein_Distance": null,
      "Chi_Square_Statistic": 66246.5298,
      "Contingency_CramerV": 0.01224,
      "Category_Coverage": 1.0,
      "Range_Coverage": null
    },
    "race": {
      "KS_Statistic": null,
      "JS_Divergence": 0.01451,
      "KL_Divergence": 0.04777,
      "Hellinger_Distance": 0.12482,
      "Total_Variation_Distance": 0.06174,
      "Wasserstein_Distance": null,
      "Chi_Square_Statistic": 17153.34585,
      "Contingency_CramerV": 0.00955,
      "Category_Coverage": 1.0,
```
33

```
      "Range_Coverage": null
    },
    "sex": {
      "KS_Statistic": null,
      "JS_Divergence": 0.00055,
      "KL_Divergence": 0.00218,
      "Hellinger_Distance": 0.02347,
      "Total_Variation_Distance": 0.03093,
      "Wasserstein_Distance": null,
      "Chi_Square_Statistic": 177.45228,
      "Contingency_CramerV": 0.00866,
      "Category_Coverage": 1.0,
      "Range_Coverage": null
    },
    "capitalgain": {
      "KS_Statistic": null,
      "JS_Divergence": 0.06878,
      "KL_Divergence": 0.24884,
      "Hellinger_Distance": 0.27313,
      "Total_Variation_Distance": 0.24636,
      "Wasserstein_Distance": null,
      "Chi_Square_Statistic": 84839.21243,
      "Contingency_CramerV": 0.00982,
      "Category_Coverage": 1.0,
      "Range_Coverage": null
    },
    "capitalloss": {
      "KS_Statistic": null,
      "JS_Divergence": 0.0004,
      "KL_Divergence": 0.00152,
      "Hellinger_Distance": 0.02007,
      "Total_Variation_Distance": 0.00877,
      "Wasserstein_Distance": null,
      "Chi_Square_Statistic": 152.11064,
      "Contingency_CramerV": 0.01112,
      "Category_Coverage": 1.0,
      "Range_Coverage": null
    },
    "hoursperweek": {
      "KS_Statistic": null,
      "JS_Divergence": 0.02164,
      "KL_Divergence": 0.08344,
      "Hellinger_Distance": 0.1478,
      "Total_Variation_Distance": 0.17799,
      "Wasserstein_Distance": null,
      "Chi_Square_Statistic": 8719.43756,
      "Contingency_CramerV": 0.00661,
      "Category_Coverage": 1.0,
```

```
      "Range_Coverage": null
    },
    "native-country": {
      "KS_Statistic": 0.4586,
      "JS_Divergence": 0.22567,
      "KL_Divergence": 0.77992,
      "Wasserstein_Distance": 1.76764,
      "Hellinger_Distance": 0.53166,
      "Total_Variation_Distance": 0.53918,
      "Range_Coverage": 1.0,
      "Chi_Square_Statistic": null,
      "Contingency_CramerV": null,
      "Category_Coverage": null
    },
    "class": {
      "KS_Statistic": null,
      "JS_Divergence": 0.00035,
      "KL_Divergence": 0.0014,
      "Hellinger_Distance": 0.01875,
      "Total_Variation_Distance": 0.02443,
      "Wasserstein_Distance": null,
      "Chi_Square_Statistic": 112.81671,
      "Contingency_CramerV": 0.00425,
      "Category_Coverage": 1.0,
      "Range_Coverage": null
    }
  }
}
```

**The complete metrics report from UC3 (.JSON file)**

```
{
  "metadata": {
    "run_id": "sdb_779c8d8d5421",
    "timestamp": "2025-12-10T17:49:49.228663",
    "real_dataset_path": "test/UC3_real.csv",
    "synthetic_dataset_path": "test/UC3_synth.csv",
    "number_of_samples_real": 109394,
    "number_of_samples_synthetic": 109394,
    "total_features": 62,
    "numerical_features": 54,
    "binary_categorical_features": 5,
    "multi_categorical_features": 3,
    "total_missing_values": 0,
    "data_completeness (%)": 100.0,
    "outliers (%)": {
      "Src Port": 19.42,
      "Dst Port": 0.0,
```

```
"Flow Duration": 20.76,
"Tot Fwd Pkts": 0.19,
"Tot Bwd Pkts": 0.19,
"TotLen Fwd Pkts": 0.3,
"TotLen Bwd Pkts": 0.0,
"Fwd Pkt Len Max": 0.0,
"Fwd Pkt Len Mean": 0.13,
"Fwd Pkt Len Std": 0.0,
"Bwd Pkt Len Max": 0.0,
"Bwd Pkt Len Mean": 0.23,
"Bwd Pkt Len Std": 0.03,
"Flow Byts/s": 14.12,
"Flow Pkts/s": 2.65,
"Flow IAT Mean": 16.67,
"Flow IAT Std": 21.15,
"Flow IAT Max": 19.07,
"Flow IAT Min": 16.1,
"Fwd IAT Tot": 14.94,
"Fwd IAT Mean": 14.7,
"Fwd IAT Std": 14.15,
"Fwd IAT Max": 14.69,
"Fwd IAT Min": 12.97,
"Bwd IAT Tot": 20.5,
"Bwd IAT Mean": 16.54,
"Bwd IAT Std": 21.06,
"Bwd IAT Max": 19.36,
"Bwd IAT Min": 18.64,
"Fwd Header Len": 0.18,
"Bwd Header Len": 0.16,
"Fwd Pkts/s": 14.19,
"Bwd Pkts/s": 2.65,
"Pkt Len Max": 0.0,
"Pkt Len Mean": 0.5,
"Pkt Len Std": 0.0,
"Pkt Len Var": 0.37,
"Pkt Size Avg": 0.48,
"Fwd Seg Size Avg": 0.13,
"Bwd Seg Size Avg": 0.23,
"Subflow Fwd Pkts": 0.19,
"Subflow Fwd Byts": 0.3,
"Subflow Bwd Pkts": 0.19,
"Subflow Bwd Byts": 0.0,
"Init Bwd Win Byts": 9.74,
"Fwd Act Data Pkts": 0.41,
"Active Mean": 13.67,
"Active Std": 0.33,
"Active Max": 13.67,
"Active Min": 13.67,
```

```
          "Idle Mean": 13.68,
          "Idle Std": 0.53,
          "Idle Max": 13.68,
          "Idle Min": 13.68
      }
  },
  "metric_definitions": {
      "Kolmogorov–Smirnov (KS) Statistic": "Measures the maximum distance between the empirical cumulative distributions of real and synthetic data for a numeric feature.",
      "Kullback-Leibler Divergence (KLD)": "Quantifies how much information is lost when approximating the real data distribution with the synthetic one. Asymmetric measure.",
      "Jensen–Shannon (JS) Divergence (JSD)": "Symmetric measure of similarity between two probability distributions derived from real and synthetic data. Lower values indicate higher similarity.",
      "Wasserstein Distance (WD)": "Quantifies the minimum 'work' required to transform one probability distribution into another, reflecting both shape and distance differences.",
      "Hellinger Distance (HD)": "Measures the distance between two probability distributions; bounded between 0 (identical) and 1 (completely dissimilar).",
      "Total Variation Distance (TVD)": "Measures the maximum absolute difference between two probability distributions. Values range from 0 (identical) to 1 (completely disjoint). Supports both numeric and categorical data.",
      "Range Coverage (RC)": "Fraction of the real data's numeric range that is covered by the synthetic data. Values close to 1 indicate the synthetic data spans the same domain as the real data.",
      "Chi-Square Statistic (CSS)": "Tests whether the observed category frequencies in the synthetic data differ significantly from those in the real data.",
      "Category Coverage (CC)": "Proportion of unique categories in the real data that also appear in the synthetic data; detects missing or underrepresented categories.",
      "Contingency Table Similarity (CV)": "Measures the strength of association between two categorical variables in real vs. synthetic datasets; used to compare inter-feature dependencies.",
      "Covariance Matrix Similarity (CMS)": "Quantifies deviation between real and synthetic covariance matrices; smaller Frobenius norm indicates closer similarity.",
      "Correlation Matrix Distance (CMD)": "Computes normalized Frobenius norm of the difference between correlation matrices; used as an overall measure of structural fidelity.",
      "Correlation Difference (Pearson) (CDP)": "Measures how much the linear (Pearson) correlations between features differ between real and synthetic datasets.",
      "Correlation Difference (Spearman) (CDS)": "Measures how much the rank (Spearman) correlations between features differ between real and synthetic datasets.",
      "Mutual Information Difference (MID)": "Captures how well nonlinear dependencies between variables are preserved; compares mutual information matrices between real and synthetic data.",
      "Centered Kernel Alignment (CKA)": "Measures similarity between real and synthetic feature representations in embedding space. Values range from 0 (no similarity) to 1 (identical representation).",
      "Average Wasserstein Embedding Distance (AWED)": "Average Wasserstein distance between real and synthetic points in embedding space. Lower values indicate better alignment of sample distributions.",
      "Neighbor Overlap (Jaccard Similarity)": "Measures how similar each sample's nearest-neighbor set is between real and synthetic data. Calculated using Jaccard index between the kNN lists of real and synthetic embeddings.",
```

```
        "Spectral Distance (SD)": "Distance between the eigenvalue spectra of real and synthetic kNN
graphs. Lower values indicate better preservation of global graph structure.",
        "Graph Structural Fidelity Score (GSFS)": "Measures global structural preservation of the kNN
graph by comparing degree distributions, clustering coefficients, and shortest-path distances. Values
range from 0 to 1, with higher values indicating better global topology preservation."
    },
    "global_metrics": {
        "Covariance_Matrix_Similarity_Frobenius": 8230321416884.392,
        "Correlation_Matrix_Distance": 0.16609,
        "Correlation_Difference_Pearson": 0.05939,
        "Correlation_Difference_Spearman": 0.3812,
        "Mutual_Information_Difference": 0.16348
    },
    "local_metrics": {
        "Src Port": {
            "KS_Statistic": 0.4691,
            "JS_Divergence": 0.19281,
            "KL_Divergence": 0.66109,
            "Wasserstein_Distance": 5729.58895,
            "Hellinger_Distance": 0.47241,
            "Total_Variation_Distance": 0.48946,
            "Range_Coverage": 1.0,
            "Chi_Square_Statistic": null,
            "Contingency_CramerV": null,
            "Category_Coverage": null
        },
        "Dst Port": {
            "KS_Statistic": 0.34778,
            "JS_Divergence": 0.15852,
            "KL_Divergence": 0.54427,
            "Wasserstein_Distance": 5042.44718,
            "Hellinger_Distance": 0.45054,
            "Total_Variation_Distance": 0.43437,
            "Range_Coverage": 1.0,
            "Chi_Square_Statistic": null,
            "Contingency_CramerV": null,
            "Category_Coverage": null
        },
        "Protocol": {
            "KS_Statistic": null,
            "JS_Divergence": 0.00038,
            "KL_Divergence": 0.00143,
            "Hellinger_Distance": 0.01989,
            "Total_Variation_Distance": 0.02404,
            "Wasserstein_Distance": null,
            "Chi_Square_Statistic": 972.17653,
            "Contingency_CramerV": 0.00304,
            "Category_Coverage": 1.0,
```

```
      "Range_Coverage": null
    },
    "Flow Duration": {
      "KS_Statistic": 0.41457,
      "JS_Divergence": 0.21901,
      "KL_Divergence": 0.84742,
      "Wasserstein_Distance": 2897412.9124,
      "Hellinger_Distance": 0.50861,
      "Total_Variation_Distance": 0.52138,
      "Range_Coverage": 1.0,
      "Chi_Square_Statistic": null,
      "Contingency_CramerV": null,
      "Category_Coverage": null
    },
    "Tot Fwd Pkts": {
      "KS_Statistic": 0.2044,
      "JS_Divergence": 0.13448,
      "KL_Divergence": 0.44138,
      "Wasserstein_Distance": 0.44744,
      "Hellinger_Distance": 0.40439,
      "Total_Variation_Distance": 0.34421,
      "Range_Coverage": 0.21429,
      "Chi_Square_Statistic": null,
      "Contingency_CramerV": null,
      "Category_Coverage": null
    },
    "Tot Bwd Pkts": {
      "KS_Statistic": 0.3074,
      "JS_Divergence": 0.22893,
      "KL_Divergence": 0.79808,
      "Wasserstein_Distance": 0.60213,
      "Hellinger_Distance": 0.5284,
      "Total_Variation_Distance": 0.53143,
      "Range_Coverage": 0.24528,
      "Chi_Square_Statistic": null,
      "Contingency_CramerV": null,
      "Category_Coverage": null
    },
    "TotLen Fwd Pkts": {
      "KS_Statistic": 0.36337,
      "JS_Divergence": 0.26163,
      "KL_Divergence": 0.90587,
      "Wasserstein_Distance": 6.40354,
      "Hellinger_Distance": 0.5895,
      "Total_Variation_Distance": 0.56954,
      "Range_Coverage": 0.12398,
      "Chi_Square_Statistic": null,
      "Contingency_CramerV": null,
```

```
      "Category_Coverage": null
    },
    "TotLen Bwd Pkts": {
      "KS_Statistic": 0.39946,
      "JS_Divergence": 0.27454,
      "KL_Divergence": 0.94346,
      "Wasserstein_Distance": 3.90614,
      "Hellinger_Distance": 0.60104,
      "Total_Variation_Distance": 0.5976,
      "Range_Coverage": 0.21639,
      "Chi_Square_Statistic": null,
      "Contingency_CramerV": null,
      "Category_Coverage": null
    },
    "Fwd Pkt Len Max": {
      "KS_Statistic": 0.40088,
      "JS_Divergence": 0.27465,
      "KL_Divergence": 0.93286,
      "Wasserstein_Distance": 3.81745,
      "Hellinger_Distance": 0.60217,
      "Total_Variation_Distance": 0.59766,
      "Range_Coverage": 0.22881,
      "Chi_Square_Statistic": null,
      "Contingency_CramerV": null,
      "Category_Coverage": null
    },
    "Fwd Pkt Len Mean": {
      "KS_Statistic": 0.43062,
      "JS_Divergence": 0.33982,
      "KL_Divergence": 1.45925,
      "Wasserstein_Distance": 1.08736,
      "Hellinger_Distance": 0.67133,
      "Total_Variation_Distance": 0.66335,
      "Range_Coverage": 0.72053,
      "Chi_Square_Statistic": null,
      "Contingency_CramerV": null,
      "Category_Coverage": null
    },
    "Fwd Pkt Len Std": {
      "KS_Statistic": 0.43396,
      "JS_Divergence": 0.32264,
      "KL_Divergence": 1.35204,
      "Wasserstein_Distance": 2.02398,
      "Hellinger_Distance": 0.64963,
      "Total_Variation_Distance": 0.65228,
      "Range_Coverage": 0.22402,
      "Chi_Square_Statistic": null,
      "Contingency_CramerV": null,
```

```json
      "Category_Coverage": null
    },
    "Bwd Pkt Len Max": {
      "KS_Statistic": 0.39947,
      "JS_Divergence": 0.27366,
      "KL_Divergence": 0.93709,
      "Wasserstein_Distance": 3.88387,
      "Hellinger_Distance": 0.60021,
      "Total_Variation_Distance": 0.59645,
      "Range_Coverage": 0.65385,
      "Chi_Square_Statistic": null,
      "Contingency_CramerV": null,
      "Category_Coverage": null
    },
    "Bwd Pkt Len Mean": {
      "KS_Statistic": 0.4223,
      "JS_Divergence": 0.32561,
      "KL_Divergence": 1.40955,
      "Wasserstein_Distance": 1.03948,
      "Hellinger_Distance": 0.64642,
      "Total_Variation_Distance": 0.65202,
      "Range_Coverage": 0.29015,
      "Chi_Square_Statistic": null,
      "Contingency_CramerV": null,
      "Category_Coverage": null
    },
    "Bwd Pkt Len Std": {
      "KS_Statistic": 0.43174,
      "JS_Divergence": 0.31912,
      "KL_Divergence": 1.37687,
      "Wasserstein_Distance": 1.95118,
      "Hellinger_Distance": 0.64446,
      "Total_Variation_Distance": 0.6465,
      "Range_Coverage": 0.59099,
      "Chi_Square_Statistic": null,
      "Contingency_CramerV": null,
      "Category_Coverage": null
    },
    "Flow Byts/s": {
      "KS_Statistic": 0.40078,
      "JS_Divergence": 0.15995,
      "KL_Divergence": 0.52912,
      "Wasserstein_Distance": 23405.98567,
      "Hellinger_Distance": 0.45734,
      "Total_Variation_Distance": 0.4139,
      "Range_Coverage": 0.064,
      "Chi_Square_Statistic": null,
      "Contingency_CramerV": null,
```

```
      "Category_Coverage": null
    },
    "Flow Pkts/s": {
      "KS_Statistic": 0.24519,
      "JS_Divergence": 0.12491,
      "KL_Divergence": 0.49297,
      "Wasserstein_Distance": 15486.06196,
      "Hellinger_Distance": 0.40401,
      "Total_Variation_Distance": 0.28762,
      "Range_Coverage": 0.2556,
      "Chi_Square_Statistic": null,
      "Contingency_CramerV": null,
      "Category_Coverage": null
    },
    "Flow IAT Mean": {
      "KS_Statistic": 0.42119,
      "JS_Divergence": 0.23122,
      "KL_Divergence": 0.90451,
      "Wasserstein_Distance": 381225.99496,
      "Hellinger_Distance": 0.52475,
      "Total_Variation_Distance": 0.53137,
      "Range_Coverage": 0.40506,
      "Chi_Square_Statistic": null,
      "Contingency_CramerV": null,
      "Category_Coverage": null
    },
    "Flow IAT Std": {
      "KS_Statistic": 0.43347,
      "JS_Divergence": 0.25765,
      "KL_Divergence": 1.25461,
      "Wasserstein_Distance": 953435.73516,
      "Hellinger_Distance": 0.56231,
      "Total_Variation_Distance": 0.55906,
      "Range_Coverage": 1.0,
      "Chi_Square_Statistic": null,
      "Contingency_CramerV": null,
      "Category_Coverage": null
    },
    "Flow IAT Max": {
      "KS_Statistic": 0.42703,
      "JS_Divergence": 0.2459,
      "KL_Divergence": 1.15093,
      "Wasserstein_Distance": 2817966.80897,
      "Hellinger_Distance": 0.54424,
      "Total_Variation_Distance": 0.5494,
      "Range_Coverage": 1.0,
      "Chi_Square_Statistic": null,
      "Contingency_CramerV": null,
```

```
          "Category_Coverage": null
        },
        "Flow IAT Min": {
          "KS_Statistic": 0.49203,
          "JS_Divergence": 0.58108,
          "KL_Divergence": 4.19784,
          "Wasserstein_Distance": 93053.68304,
          "Hellinger_Distance": 0.87989,
          "Total_Variation_Distance": 0.90233,
          "Range_Coverage": 0.01468,
          "Chi_Square_Statistic": null,
          "Contingency_CramerV": null,
          "Category_Coverage": null
        },
        "Fwd IAT Tot": {
          "KS_Statistic": 0.40425,
          "JS_Divergence": 0.17118,
          "KL_Divergence": 0.67046,
          "Wasserstein_Distance": 139788.20406,
          "Hellinger_Distance": 0.46762,
          "Total_Variation_Distance": 0.40757,
          "Range_Coverage": 1.0,
          "Chi_Square_Statistic": null,
          "Contingency_CramerV": null,
          "Category_Coverage": null
        },
        "Fwd IAT Mean": {
          "KS_Statistic": 0.41091,
          "JS_Divergence": 0.16247,
          "KL_Divergence": 0.64807,
          "Wasserstein_Distance": 32883.31645,
          "Hellinger_Distance": 0.45287,
          "Total_Variation_Distance": 0.41518,
          "Range_Coverage": 1.0,
          "Chi_Square_Statistic": null,
          "Contingency_CramerV": null,
          "Category_Coverage": null
        },
        "Fwd IAT Std": {
          "KS_Statistic": 0.42162,
          "JS_Divergence": 0.17264,
          "KL_Divergence": 0.66579,
          "Wasserstein_Distance": 52039.90479,
          "Hellinger_Distance": 0.47076,
          "Total_Variation_Distance": 0.42639,
          "Range_Coverage": 1.0,
          "Chi_Square_Statistic": null,
          "Contingency_CramerV": null,
```

```
      "Category_Coverage": null
    },
    "Fwd IAT Max": {
      "KS_Statistic": 0.41243,
      "JS_Divergence": 0.1743,
      "KL_Divergence": 0.6925,
      "Wasserstein_Distance": 96811.37435,
      "Hellinger_Distance": 0.4741,
      "Total_Variation_Distance": 0.41824,
      "Range_Coverage": 1.0,
      "Chi_Square_Statistic": null,
      "Contingency_CramerV": null,
      "Category_Coverage": null
    },
    "Fwd IAT Min": {
      "KS_Statistic": 0.3799,
      "JS_Divergence": 0.1248,
      "KL_Divergence": 0.51064,
      "Wasserstein_Distance": 9753.43155,
      "Hellinger_Distance": 0.36984,
      "Total_Variation_Distance": 0.38936,
      "Range_Coverage": 1.0,
      "Chi_Square_Statistic": null,
      "Contingency_CramerV": null,
      "Category_Coverage": null
    },
    "Bwd IAT Tot": {
      "KS_Statistic": 0.41243,
      "JS_Divergence": 0.23206,
      "KL_Divergence": 1.02963,
      "Wasserstein_Distance": 2867321.19071,
      "Hellinger_Distance": 0.5273,
      "Total_Variation_Distance": 0.5311,
      "Range_Coverage": 1.0,
      "Chi_Square_Statistic": null,
      "Contingency_CramerV": null,
      "Category_Coverage": null
    },
    "Bwd IAT Mean": {
      "KS_Statistic": 0.41635,
      "JS_Divergence": 0.25083,
      "KL_Divergence": 1.25869,
      "Wasserstein_Distance": 512625.69106,
      "Hellinger_Distance": 0.55482,
      "Total_Variation_Distance": 0.54098,
      "Range_Coverage": 0.7101,
      "Chi_Square_Statistic": null,
      "Contingency_CramerV": null,
```

```
      "Category_Coverage": null
    },
    "Bwd IAT Std": {
      "KS_Statistic": 0.42395,
      "JS_Divergence": 0.25253,
      "KL_Divergence": 1.25333,
      "Wasserstein_Distance": 1147340.62745,
      "Hellinger_Distance": 0.55709,
      "Total_Variation_Distance": 0.55142,
      "Range_Coverage": 1.0,
      "Chi_Square_Statistic": null,
      "Contingency_CramerV": null,
      "Category_Coverage": null
    },
    "Bwd IAT Max": {
      "KS_Statistic": 0.42092,
      "JS_Divergence": 0.24478,
      "KL_Divergence": 1.20843,
      "Wasserstein_Distance": 2797807.7622,
      "Hellinger_Distance": 0.54398,
      "Total_Variation_Distance": 0.54369,
      "Range_Coverage": 1.0,
      "Chi_Square_Statistic": null,
      "Contingency_CramerV": null,
      "Category_Coverage": null
    },
    "Bwd IAT Min": {
      "KS_Statistic": 0.4036,
      "JS_Divergence": 0.43567,
      "KL_Divergence": 3.11815,
      "Wasserstein_Distance": 3656.16947,
      "Hellinger_Distance": 0.74223,
      "Total_Variation_Distance": 0.79115,
      "Range_Coverage": 0.2216,
      "Chi_Square_Statistic": null,
      "Contingency_CramerV": null,
      "Category_Coverage": null
    },
    "Bwd PSH Flags": {
      "KS_Statistic": null,
      "JS_Divergence": 0.01143,
      "KL_Divergence": 0.03528,
      "Hellinger_Distance": 0.11327,
      "Total_Variation_Distance": 0.04237,
      "Wasserstein_Distance": null,
      "Chi_Square_Statistic": 67332.03822,
      "Contingency_CramerV": 0.00041,
      "Category_Coverage": 1.0,
```

```
          "Range_Coverage": null
      },
      "Fwd Header Len": {
          "KS_Statistic": 0.41088,
          "JS_Divergence": 0.25423,
          "KL_Divergence": 1.03152,
          "Wasserstein_Distance": 17.06179,
          "Hellinger_Distance": 0.58138,
          "Total_Variation_Distance": 0.54568,
          "Range_Coverage": 0.19955,
          "Chi_Square_Statistic": null,
          "Contingency_CramerV": null,
          "Category_Coverage": null
      },
      "Bwd Header Len": {
          "KS_Statistic": 0.30687,
          "JS_Divergence": 0.27328,
          "KL_Divergence": 1.10176,
          "Wasserstein_Distance": 13.11039,
          "Hellinger_Distance": 0.59484,
          "Total_Variation_Distance": 0.57955,
          "Range_Coverage": 0.30703,
          "Chi_Square_Statistic": null,
          "Contingency_CramerV": null,
          "Category_Coverage": null
      },
      "Fwd Pkts/s": {
          "KS_Statistic": 0.40663,
          "JS_Divergence": 0.18641,
          "KL_Divergence": 0.63413,
          "Wasserstein_Distance": 12038.17341,
          "Hellinger_Distance": 0.49761,
          "Total_Variation_Distance": 0.44974,
          "Range_Coverage": 0.1358,
          "Chi_Square_Statistic": null,
          "Contingency_CramerV": null,
          "Category_Coverage": null
      },
      "Bwd Pkts/s": {
          "KS_Statistic": 0.21523,
          "JS_Divergence": 0.12675,
          "KL_Divergence": 0.52392,
          "Wasserstein_Distance": 13129.65275,
          "Hellinger_Distance": 0.40706,
          "Total_Variation_Distance": 0.31142,
          "Range_Coverage": 0.49485,
          "Chi_Square_Statistic": null,
          "Contingency_CramerV": null,
```

```json
      "Category_Coverage": null
    },
    "Pkt Len Max": {
      "KS_Statistic": 0.398,
      "JS_Divergence": 0.27525,
      "KL_Divergence": 0.93819,
      "Wasserstein_Distance": 3.89044,
      "Hellinger_Distance": 0.6032,
      "Total_Variation_Distance": 0.59817,
      "Range_Coverage": 0.2161,
      "Chi_Square_Statistic": null,
      "Contingency_CramerV": null,
      "Category_Coverage": null
    },
    "Pkt Len Mean": {
      "KS_Statistic": 0.42203,
      "JS_Divergence": 0.32324,
      "KL_Divergence": 1.30959,
      "Wasserstein_Distance": 0.95121,
      "Hellinger_Distance": 0.64962,
      "Total_Variation_Distance": 0.6509,
      "Range_Coverage": 0.34354,
      "Chi_Square_Statistic": null,
      "Contingency_CramerV": null,
      "Category_Coverage": null
    },
    "Pkt Len Std": {
      "KS_Statistic": 0.42913,
      "JS_Divergence": 0.30638,
      "KL_Divergence": 1.14447,
      "Wasserstein_Distance": 1.76264,
      "Hellinger_Distance": 0.62929,
      "Total_Variation_Distance": 0.64013,
      "Range_Coverage": 0.36018,
      "Chi_Square_Statistic": null,
      "Contingency_CramerV": null,
      "Category_Coverage": null
    },
    "Pkt Len Var": {
      "KS_Statistic": 0.4211,
      "JS_Divergence": 0.23974,
      "KL_Divergence": 1.01212,
      "Wasserstein_Distance": 24.29068,
      "Hellinger_Distance": 0.55819,
      "Total_Variation_Distance": 0.50959,
      "Range_Coverage": 0.12269,
      "Chi_Square_Statistic": null,
      "Contingency_CramerV": null,
```

```json
      "Category_Coverage": null
    },
    "FIN Flag Cnt": {
      "KS_Statistic": null,
      "JS_Divergence": 0.00108,
      "KL_Divergence": 0.0044,
      "Hellinger_Distance": 0.03287,
      "Total_Variation_Distance": 0.04019,
      "Wasserstein_Distance": null,
      "Chi_Square_Statistic": 897.78534,
      "Contingency_CramerV": 0.00102,
      "Category_Coverage": 1.0,
      "Range_Coverage": null
    },
    "SYN Flag Cnt": {
      "KS_Statistic": null,
      "JS_Divergence": 0.00344,
      "KL_Divergence": 0.01331,
      "Hellinger_Distance": 0.05871,
      "Total_Variation_Distance": 0.06938,
      "Wasserstein_Distance": null,
      "Chi_Square_Statistic": 3392.25381,
      "Contingency_CramerV": 0.00112,
      "Category_Coverage": 1.0,
      "Range_Coverage": null
    },
    "PSH Flag Cnt": {
      "KS_Statistic": null,
      "JS_Divergence": 0.01157,
      "KL_Divergence": 0.03572,
      "Hellinger_Distance": 0.11402,
      "Total_Variation_Distance": 0.04282,
      "Wasserstein_Distance": null,
      "Chi_Square_Statistic": 68763.19652,
      "Contingency_CramerV": 0.00133,
      "Category_Coverage": 1.0,
      "Range_Coverage": null
    },
    "ACK Flag Cnt": {
      "KS_Statistic": null,
      "JS_Divergence": 0.00098,
      "KL_Divergence": 0.00401,
      "Hellinger_Distance": 0.03138,
      "Total_Variation_Distance": 0.03859,
      "Wasserstein_Distance": null,
      "Chi_Square_Statistic": 821.00382,
      "Contingency_CramerV": 0.00125,
      "Category_Coverage": 1.0,
```

```
      "Range_Coverage": null
    },
    "Down/Up Ratio": {
      "KS_Statistic": null,
      "JS_Divergence": 0.0244,
      "KL_Divergence": 0.084,
      "Hellinger_Distance": 0.17309,
      "Total_Variation_Distance": 0.08643,
      "Wasserstein_Distance": null,
      "Chi_Square_Statistic": 3488519.55539,
      "Contingency_CramerV": 0.00494,
      "Category_Coverage": 1.0,
      "Range_Coverage": null
    },
    "Pkt Size Avg": {
      "KS_Statistic": 0.41735,
      "JS_Divergence": 0.31782,
      "KL_Divergence": 1.28912,
      "Wasserstein_Distance": 1.10235,
      "Hellinger_Distance": 0.64345,
      "Total_Variation_Distance": 0.64416,
      "Range_Coverage": 0.26102,
      "Chi_Square_Statistic": null,
      "Contingency_CramerV": null,
      "Category_Coverage": null
    },
    "Fwd Seg Size Avg": {
      "KS_Statistic": 0.43093,
      "JS_Divergence": 0.34003,
      "KL_Divergence": 1.46052,
      "Wasserstein_Distance": 1.08679,
      "Hellinger_Distance": 0.67156,
      "Total_Variation_Distance": 0.66338,
      "Range_Coverage": 0.73224,
      "Chi_Square_Statistic": null,
      "Contingency_CramerV": null,
      "Category_Coverage": null
    },
    "Bwd Seg Size Avg": {
      "KS_Statistic": 0.42249,
      "JS_Divergence": 0.32625,
      "KL_Divergence": 1.40929,
      "Wasserstein_Distance": 1.03987,
      "Hellinger_Distance": 0.64705,
      "Total_Variation_Distance": 0.65261,
      "Range_Coverage": 0.29198,
      "Chi_Square_Statistic": null,
      "Contingency_CramerV": null,
```

```json
      "Category_Coverage": null
    },
    "Subflow Fwd Pkts": {
      "KS_Statistic": 0.20459,
      "JS_Divergence": 0.13521,
      "KL_Divergence": 0.44389,
      "Wasserstein_Distance": 0.44877,
      "Hellinger_Distance": 0.4055,
      "Total_Variation_Distance": 0.34573,
      "Range_Coverage": 0.22619,
      "Chi_Square_Statistic": null,
      "Contingency_CramerV": null,
      "Category_Coverage": null
    },
    "Subflow Fwd Byts": {
      "KS_Statistic": 0.36323,
      "JS_Divergence": 0.26155,
      "KL_Divergence": 0.90476,
      "Wasserstein_Distance": 6.40431,
      "Hellinger_Distance": 0.58941,
      "Total_Variation_Distance": 0.56936,
      "Range_Coverage": 0.12358,
      "Chi_Square_Statistic": null,
      "Contingency_CramerV": null,
      "Category_Coverage": null
    },
    "Subflow Bwd Pkts": {
      "KS_Statistic": 0.30722,
      "JS_Divergence": 0.22897,
      "KL_Divergence": 0.79768,
      "Wasserstein_Distance": 0.60444,
      "Hellinger_Distance": 0.52834,
      "Total_Variation_Distance": 0.53191,
      "Range_Coverage": 0.24528,
      "Chi_Square_Statistic": null,
      "Contingency_CramerV": null,
      "Category_Coverage": null
    },
    "Subflow Bwd Byts": {
      "KS_Statistic": 0.39945,
      "JS_Divergence": 0.27419,
      "KL_Divergence": 0.94404,
      "Wasserstein_Distance": 3.90628,
      "Hellinger_Distance": 0.60072,
      "Total_Variation_Distance": 0.59709,
      "Range_Coverage": 0.21967,
      "Chi_Square_Statistic": null,
      "Contingency_CramerV": null,
```

```json
      "Category_Coverage": null
    },
    "Init Bwd Win Byts": {
      "KS_Statistic": 0.36941,
      "JS_Divergence": 0.59987,
      "KL_Divergence": 4.1286,
      "Wasserstein_Distance": 241.86987,
      "Hellinger_Distance": 0.90016,
      "Total_Variation_Distance": 0.93579,
      "Range_Coverage": 1.0,
      "Chi_Square_Statistic": null,
      "Contingency_CramerV": null,
      "Category_Coverage": null
    },
    "Fwd Act Data Pkts": {
      "KS_Statistic": 0.08071,
      "JS_Divergence": 0.00023,
      "KL_Divergence": 0.00435,
      "Wasserstein_Distance": 0.11182,
      "Hellinger_Distance": 0.01697,
      "Total_Variation_Distance": 0.00119,
      "Range_Coverage": 0.17073,
      "Chi_Square_Statistic": null,
      "Contingency_CramerV": null,
      "Category_Coverage": null
    },
    "Active Mean": {
      "KS_Statistic": 0.46533,
      "JS_Divergence": 0.19863,
      "KL_Divergence": 0.68006,
      "Wasserstein_Distance": 71423.44023,
      "Hellinger_Distance": 0.50708,
      "Total_Variation_Distance": 0.48675,
      "Range_Coverage": 1.0,
      "Chi_Square_Statistic": null,
      "Contingency_CramerV": null,
      "Category_Coverage": null
    },
    "Active Std": {
      "KS_Statistic": 0.4968,
      "JS_Divergence": 0.21477,
      "KL_Divergence": 0.70911,
      "Wasserstein_Distance": 2099.6328,
      "Hellinger_Distance": 0.53933,
      "Total_Variation_Distance": 0.49869,
      "Range_Coverage": 1.0,
      "Chi_Square_Statistic": null,
      "Contingency_CramerV": null,
```

```
          "Category_Coverage": null
        },
        "Active Max": {
          "KS_Statistic": 0.46609,
          "JS_Divergence": 0.19998,
          "KL_Divergence": 0.68957,
          "Wasserstein_Distance": 73293.1477,
          "Hellinger_Distance": 0.50876,
          "Total_Variation_Distance": 0.48775,
          "Range_Coverage": 1.0,
          "Chi_Square_Statistic": null,
          "Contingency_CramerV": null,
          "Category_Coverage": null
        },
        "Active Min": {
          "KS_Statistic": 0.46674,
          "JS_Divergence": 0.19925,
          "KL_Divergence": 0.67835,
          "Wasserstein_Distance": 72438.67488,
          "Hellinger_Distance": 0.50777,
          "Total_Variation_Distance": 0.48724,
          "Range_Coverage": 1.0,
          "Chi_Square_Statistic": null,
          "Contingency_CramerV": null,
          "Category_Coverage": null
        },
        "Idle Mean": {
          "KS_Statistic": 0.46104,
          "JS_Divergence": 0.27023,
          "KL_Divergence": 1.2781,
          "Wasserstein_Distance": 2787190.1759,
          "Hellinger_Distance": 0.58909,
          "Total_Variation_Distance": 0.57535,
          "Range_Coverage": 1.0,
          "Chi_Square_Statistic": null,
          "Contingency_CramerV": null,
          "Category_Coverage": null
        },
        "Idle Std": {
          "KS_Statistic": 0.50405,
          "JS_Divergence": 0.21816,
          "KL_Divergence": 0.72001,
          "Wasserstein_Distance": 56194.34943,
          "Hellinger_Distance": 0.54049,
          "Total_Variation_Distance": 0.50535,
          "Range_Coverage": 1.0,
          "Chi_Square_Statistic": null,
          "Contingency_CramerV": null,
```

```json
        "Category_Coverage": null
      },
      "Idle Max": {
        "KS_Statistic": 0.46037,
        "JS_Divergence": 0.26884,
        "KL_Divergence": 1.25564,
        "Wasserstein_Distance": 2813684.12128,
        "Hellinger_Distance": 0.58686,
        "Total_Variation_Distance": 0.57416,
        "Range_Coverage": 1.0,
        "Chi_Square_Statistic": null,
        "Contingency_CramerV": null,
        "Category_Coverage": null
      },
      "Idle Min": {
        "KS_Statistic": 0.46115,
        "JS_Divergence": 0.27116,
        "KL_Divergence": 1.28758,
        "Wasserstein_Distance": 2764801.00401,
        "Hellinger_Distance": 0.59064,
        "Total_Variation_Distance": 0.57579,
        "Range_Coverage": 1.0,
        "Chi_Square_Statistic": null,
        "Contingency_CramerV": null,
        "Category_Coverage": null
      },
      "Label": {
        "KS_Statistic": null,
        "JS_Divergence": 0.19708,
        "KL_Divergence": 0.65169,
        "Hellinger_Distance": 0.48131,
        "Total_Variation_Distance": 0.49763,
        "Wasserstein_Distance": null,
        "Chi_Square_Statistic": 6432991.4225,
        "Contingency_CramerV": 0.00557,
        "Category_Coverage": 1.0,
        "Range_Coverage": null
      }
    }
}
```